\documentclass[11pt]{article}
\usepackage{coling2020}
\usepackage{times}
\usepackage{url}
\usepackage{latexsym}

\colingfinalcopy % Uncomment this line for the final submission

\usepackage{booktabs}
\usepackage{multirow}
\usepackage{adjustbox}
\usepackage{tikz-dependency}
\newcommand{\stddev}[2]{\ensuremath{#1_{\color{darkgray}{\pm #2}}}} 

\usepackage{color, colortbl}
\definecolor{Gray}{gray}{0.9}

\usepackage{microtype}
\usepackage{tabto}

\usepackage{amsthm}
\usepackage{amsfonts}
\usepackage{amssymb}

  \newtheoremstyle{mine}% <name>
  {3pt}% <Space above>
  {3pt}% <Space below>
  {\bfshape}% <Body font>
  {}% <Indent amount>
  {\bfshape}% <Theorem head font>
  {:}% <Punctuation after theorem head>
  {.5em}% <Space after theorem headi>
  {}% <Theorem head spec (can be left empty, meaning `normal')>

\theoremstyle{definition}
\newtheorem{mydef}{Definition}

\newcommand{\concept}[1]{\textbf{{#1}}}

\renewcommand{\vec}[1]{\mathbf{#1}}

\title{A Unifying Theory of Transition-based and Sequence Labeling Parsing}

\author{Carlos G\'omez-Rodr\'iguez \qquad  Michalina Strzyz \qquad David Vilares\\
  Universidade da Coru\~na, CITIC \\
  FASTPARSE Lab, LyS Research Group \\
    Departamento de Ciencias de la Computación y Tecnologías de la Información\\
  Campus de Elvi\~na, s/n, 15071 A Coru\~na, Spain\\
  {\tt \{carlos.gomez,michalina.strzyz,david.vilares\}@udc.es}}

\date{}

\begin{document}
\maketitle
\begin{abstract}
We define a mapping from transition-based parsing algorithms that
read sentences from left to right
to sequence labeling encodings of syntactic trees.
This not only establishes a theoretical relation between transition-based parsing and sequence-labeling parsing, but also provides a method to obtain new encodings for fast and simple sequence labeling parsing from the many existing transition-based parsers for different  
formalisms. Applying it to dependency parsing, we implement sequence labeling versions of four algorithms, showing that they are learnable and obtain comparable performance to existing encodings.
\end{abstract}

\blfootnote{This work is licensed under a Creative Commons Attribution 4.0 International Licence. Licence details: \url{http://creativecommons.org/licenses/by/4.0/}.}

\section{Introduction}
\label{intro}

Transition-based parsing algorithms compute the syntax or semantics of sentences as graphs;
for instance in the form of dependency \cite{fraser1989parsing,yamada-matsumoto-2003-statistical,nivre-etal-2004-memory}, phrase-structure \cite{sagae-lavie-2005-classifier,zhang-clark-2009-transition,zhu-etal-2013-fast} or meaning representations \cite{wang-etal-2015-transition,swayamdipta-etal-2016-greedy,hershcovich-etal-2018-multitask}. 

These algorithms 
define an abstract state machine where each state (configuration) holds a structured representation, as well as auxiliary data structures (often, but not always, a buffer and a stack of tokens). Shift-reduce actions (transitions) are defined to move the system between states until a full parse is found.
The transition to take at each state was traditionally predicted by data-driven classifiers based on local decisions and rich feature representations \cite{zhang-nivre-2011-transition}. With the adoption of deep learning, which can globally contextualize word representations, the dependency on hand-crafted features has been drastically reduced \cite{kiperwasser-goldberg-2016-simple,shi-etal-2017-fast}; and it has also been shown that alternative ways to attack the problem can be practical. 

More particularly, several parsing problems have been cast as a machine translation task, where a sequence-to-sequence (seq2seq) network maps the sentence into a string of arbitrary length that encodes a linearized graph \cite{Vinyals2015,li-etal-2018-seq2seq,konstas-etal-2017-neural}. To a certain extent, the attention mechanism in these seq2seq models can be seen as an abstraction of the stack and buffer in transition-based parsers, where the attention weights mark the relevant words to generate the next component of the output string. 
Alternatively, some authors have reduced constituent and dependency parsing to sequence labeling, where given an input sentence of length $n$, the output has length $n$ too, assigning one label to each word 
\cite{gomez-rodriguez-vilares-2018-constituent,strzyz-etal-2019-viable}. However, these reductions have consisted in defining custom encodings for the output structure, which cannot be automatically derived.

In this context, some studies have linked transition-based parsers to seq2seq architectures,
as in \cite{li-etal-2018-seq2seq}, but to the best of our knowledge there is no unified framework or theory for transition-based and sequence labeling parsing.

\paragraph{Contribution} (i) Our first contribution is theoretical, connecting the transition-based and sequence labeling parsing paradigms. We give a broad definition of a left-to-right transition system, covering the majority of transition-based parsers, and prove that the transitions produced by such systems for a sentence of length $n$ can be mapped to a sequence of $n$ labels, hence providing a mapping from transition-based parsers to sequence labeling parsers. (ii) The second contribution is empirical, applied to dependency parsing. We implement projective and non-projective transition-based algorithms, cast them in a sequence labeling setup and show that they are learnable, 
and even outperform some existing custom encodings for parsing as labeling.
The source code is available at \url{https://github.com/mstrise/dep2label}.

\begin{figure*}[t]
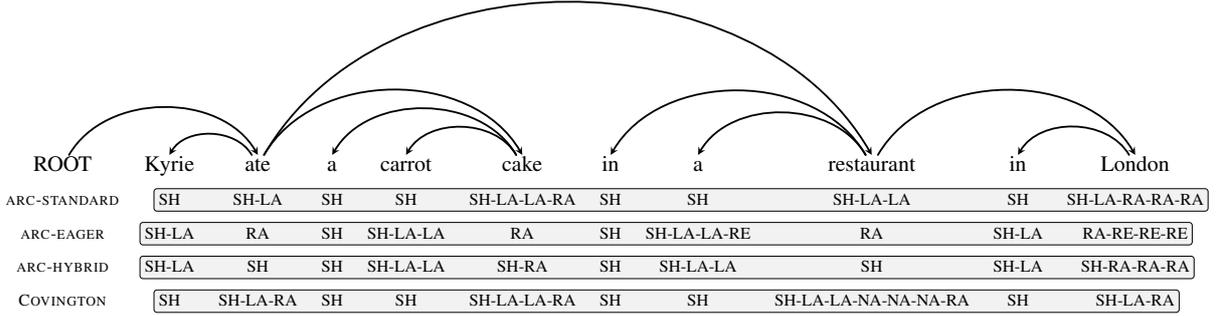
{}
\centering
\begin{adjustbox}{width=1\textwidth}
\begin{dependency}[hide label, theme=simple, arc angle=60, very thick]
\begin{deptext}[column sep=1em]
  \Large{ROOT} \& \Large{Kyrie} \& \Large{ate} \& \Large{a} \& \Large{carrot} \& \Large{cake} \& \Large{in} \& \Large{a} \& \Large{restaurant} \& \Large{in} \& \Large{London} \\
  \\
  \textsc{arc-standard}\&SH\&SH-LA\&SH\&SH\&SH-LA-LA-RA\&SH\&SH\&SH-LA-LA\&SH\&SH-LA-RA-RA-RA \\ \\
  \textsc{arc-eager}\&SH-LA\&RA\&SH\&SH-LA-LA\&RA\&SH\&SH-LA-LA-RE\&RA\&SH-LA\&RA-RE-RE-RE \\ \\
  \textsc{arc-hybrid}\&SH-LA\&SH\&SH\&SH-LA-LA\&SH-RA\&SH\&SH-LA-LA\&SH\&SH-LA\&SH-RA-RA-RA \\ \\
  \textsc{Covington}\&SH\&SH-LA-RA\&SH\&SH\&SH-LA-LA-RA\&SH\&SH\&SH-LA-LA-NA-NA-NA-RA\&SH\&SH-LA-RA \\ 
 
  \\
\end{deptext}
\depedge{1}{3}{}
\depedge{3}{2}{}
\depedge{6}{4}{}
\depedge{6}{5}{}
\depedge{3}{6}{}
\depedge{9}{7}{}
\depedge{9}{8}{}
\depedge{3}{9}{}
\depedge{11}{10}{}
\depedge{9}{11}{}

\wordgroup[group style={fill=gray!10, draw=black}]{3}{11}{2}{noun2}
\wordgroup[group style={fill=gray!10, draw=black}]{5}{11}{2}{noun2}
\wordgroup[group style={fill=gray!10, draw=black}]{7}{11}{2}{noun2}
\wordgroup[group style={fill=gray!10, draw=black}]{9}{11}{2}{noun2}
\end{dependency}
\end{adjustbox}
\caption{Examples of label distribution based on the sequence of actions for the corresponding transition systems. \textsc{la}: Left Arc, \textsc{ra}: Right Arc, \textsc{sh}: Shift, \textsc{na}: No-Arc. 
The last shift action in the sequence (which just terminates the computation and is always present) is omitted in the Covington system.}
\label{fig:encodings}
\end{figure*}

\section{Preliminaries}

\label{sec:preliminaries}

Let $\vec{w} = w_1 \ldots w_n$ be an input sentence. We will denote by $\mathbb{P}_{\vec{w}}$ the set of possible well-formed parses for ${\vec{w}}$ in the relevant parsing formalism (e.g. in projective dependency parsing, $\mathbb{P}_{\vec{w}}$ is the set of projective dependency trees of $n$ nodes).

Following \newcite{nivre-2008-algorithms}, with some adaptations to generalize the notions beyond dependency parsing, we can define a \concept{transition system} as a quadruple $S = (C,T,c_s,C_t)$ where:
\begin{itemize}
    \item $C$ is a set of configurations, such that each configuration $c \in C$ contains at least a partially-built parse $P_c$ (be it a partial syntactic dependency tree, semantic dependency tree, constituent tree, or any other structure that can be built using the input words), in addition to any other data structures needed by each specific transition system,
    \item $T$ is a set of transitions, i.e. partial functions $t: C \rightarrow C$,
    \item $c_s$ is an initialization function, mapping an input sentence $\vec{w} = w_1 \ldots w_n$ to an initial configuration in $C$,
    \item $C_t \subseteq C$ is a set of terminal configurations.
\end{itemize}

A transition system parses an input sentence $\vec{w}$ by starting in the initial configuration $c_s(\vec{w})$, and applying a sequence of transitions until a final configuration $c_f \in C_t$ is reached. At that point, the parse $P_{c_f} \in \mathbb{P}_{\vec{w}}$ is returned.

Thus, for a transition system $S = (C,T,c_s,C_t)$, we can define a \concept{computation} of $S$ on $\vec{w}$
as a sequence of configurations $c_0, c_1, \ldots, c_m$ such that each $c_{i}$ is obtained from $c_{i-1}$ by applying a transition $t_i$. Such a computation is \concept{complete} for $\vec{w}$ if $c_0 = c_s(\vec{w})$ and $c_m \in C_t$. We denote by $\mathbb{C}^S_{\vec{w}}$ the set of complete computations of $S$ on $\vec{w}$. 
Note that a computation is uniquely determined by its starting configuration and the sequence of transitions $t_1,\ldots,t_m$ applied to it. A complete computation is uniquely determined by the sequence of transitions $t_1,\ldots,t_m$, as the initial configuration is fixed.

Finally, we define a \concept{static oracle} for a transition system $S = (C,T,c_s,C_t)$ as a function $\omega: \mathbb{P}_{\vec{w}} \rightarrow \mathbb{C}^S_{\vec{w}}$
such that for every $P \in \mathbb{P}_{\vec{w}}$, the parse associated with the final configuration in $\omega(P)$ is $P$. That is, given a gold parse for the sentence $\vec{w}$, a static oracle returns a (canonical) computation of $S$ that produces that gold parse on $\vec{w}$. Note that a correct transition system should be able to produce all the possible well-formed parses in $\mathbb{P}_{\vec{w}}$, and hence, a static oracle must exist.\footnote{The traditional definition of a static oracle was as a function that returns one transition at a time, so that the computation is obtained by applying the oracle repeatedly until a final configuration is reached. However, as argued by \newcite{goldberg-nivre-2012-dynamic}, such functions only apply to transitions in the canonical transition sequence defined by the oracle, so static oracles are only correct as functions from gold parses to computations as considered here.}

\section{Mapping transition-based parsers to sequence labeling parsers}

\label{sec:mapping}

We first formally define our mapping from transition systems to sequence labeling parsers, to then explain it in detail, provide examples, and analyze how it applies to different transition systems:

\begin{mydef}
\label{def:lefttoright}
Let $S = (C,T,c_s,C_t)$ be a transition system. We say that $S$ is a \concept{left-to-right transition system} if there is a subset of transitions $T_s \subseteq T$, called the \concept{read transitions}, which satisfy the following conditions:
\begin{enumerate}
    \item Every sequence of transitions $t_1,\ldots,t_m$ corresponding to a complete computation for a sentence $w_1 \ldots w_n$ has exactly $n$ read transitions, one of which is $t_1$.
    \item There is a constant value $k$ such that, for each $1 \le i \le n$, the partial parse contained in a computation starting at $c_s$ and containing $i$ read transitions in its transition sequence is a partial parse over the substring $w_1 \ldots w_{i+k}$.
\end{enumerate}
\end{mydef}

Such parsers can be mapped into a sequence labeling encoding as follows:

\begin{mydef}
Let $S = (C,T,c_s,C_t)$ be a left-to-right transition system. Let $\gamma = c_0, \ldots, c_m$ be a complete computation of $S$ on a sentence $w_1 \ldots w_n$. By the first condition of a left-to-right transition system, we know that the sequence of transitions of $\gamma$ is of the form 

\[t^r_1, t^1_1, \ldots, t^{m_1}_1, t^r_2, t^1_2, \ldots, t^{m_2}_2, \ldots, t^r_n, t^1_n, \ldots, t^{m_n}_n\]

where each $t^r_i$ is a read transition, and $t^j_i$ is the $j$th consecutive non-read transition after $t^r_i$.

Then, we define the \concept{label sequence} associated to $\gamma$, denoted $L(\gamma)$, as the sequence of $n$ labels where the $i$th label is $t^r_i, t^1_i, \ldots, t^{m_i}_i$.
\end{mydef}

Thus, informally speaking, the notion of a transition system being left-to-right corresponds to the presence of a transition (or set of transitions) that read a new word from the input, in left-to-right order. In transition systems where such transitions are present, processing a sentence of length $n$ requires $n$ of them (as each of the $n$ words must be read from the input); and we can use this property to split transition sequences for full trees into $n$ labels (one per word), each led by a read transition.
The constant $k$ in Definition \ref{def:lefttoright} is an offset between the number of read transitions that have been executed at each given state and the words that can appear in parses at that state, as seen in the examples below.

Thus, the above definition provides a mapping from complete computations of a transition system to a label sequence. Since, as explained in Section \ref{sec:preliminaries}, every correct transition system has at least one static oracle that maps gold parses to complete computations, it is easy to compose both mappings to define a mapping from left-to-right transition systems to sequence labeling parsers, where labels are subsequences of transitions:

\begin{mydef}
Let $S = (C,T,c_s,C_t)$ be a left-to-right transition system, $\vec{w} = w_1 \ldots w_n$ a sentence of length $n$, and $\omega$ a static oracle for $S$. Then, we define the \concept{sequence labeling encoding} associated to $S$ as the mapping $\kappa : \mathbb{P}_{\vec{w}} \rightarrow (T^*)^n$ such that for each $P \in \mathbb{P}_{\vec{w}}$, $\kappa(P) = L(\omega(P))$.
\end{mydef}

\begin{table}[]
\centering
\small{
\begin{adjustbox}{max width=1\columnwidth}
\renewcommand{\arraystretch}{1}
\begin{tabular}{@{}l|l@{}}
\toprule
\multicolumn{2}{c}{Transition systems} \\ \midrule
Initial config $c_s(w_1 \ldots w_n)$: $([root],[1...n],\emptyset)$ &Initial config $c_s(w_1 \ldots w_n)$: $( [],[],[root,1 ... n], \emptyset )$ \\
Final configs: $C_t = \{([],[],P)\}$ & Final configs: $C_t = \{( \lambda_1,\lambda_2,[], P )\}$ \\ \hline
\textsc{arc-standard}\mbox{} & \textsc{covington non-projective}\mbox{} \\ \hline
\textsc{sh} $(\sigma, b|\beta, P) \Rightarrow (\sigma|b, \beta, P)$ & \textsc{sh}$_{c}$ $( \lambda_1, \lambda_2, b|\beta,P ) \Rightarrow ( \lambda_1 \cdot \lambda_2|b, [], \beta, P )$  \\
\textsc{la}$_{st}$ $(\sigma|s_1|s_0, \beta, P)\Rightarrow(\sigma|s_0, \beta, P\cup\{s_0\rightarrow s_1\})$ & \textsc{no-arc}$_{c}$ $( \lambda_1|s, \lambda_2, \beta,P ) \Rightarrow ( \lambda_1, s|\lambda_2, \beta, P )$ \\
\textsc{ra}$_{st}$ $(\sigma|s_1|s_0, \beta, P) \Rightarrow (\sigma|s_1, \beta, P\cup\{s_1\rightarrow s_0\}) $ & \textsc{la}$_{c}$ $( \lambda_1|s, \lambda_2, b|\beta,P ) \Rightarrow ( \lambda_1, s|\lambda_2,b|\beta, P \cup \{ b \rightarrow s \} )$ 
\\ \cline{1-1} 
\textsc{arc-eager}\mbox{} & \qquad \qquad if $\nexists k: k \rightarrow s \in P$ (single head)\\\cline{1-1}
\textsc{sh} $(\sigma, b|\beta, P) \Rightarrow (\sigma|b, \beta, P)$ & \qquad \qquad if $\nexists$ path $s \rightarrow ... \rightarrow b$ in $P$ (acyclicity) \\
\textsc{la}$_{e}$ $(\sigma|s, b|\beta, P) \Rightarrow (\sigma, b|\beta, P\cup\{b\rightarrow s\})$ & \textsc{ra}$_{c}$ $( \lambda_1|s, \lambda_2, b|\beta,P ) \Rightarrow ( \lambda_1, s|\lambda_2,b|\beta, P \cup \{ s \rightarrow b \} )$\\
\qquad \qquad if $\nexists k: k \rightarrow s \in P$ & \qquad \qquad if $\nexists k: k \rightarrow b \in P$ (single head) \\
\textsc{ra}$_{e}$ $(\sigma|s, b|\beta, P) \Rightarrow (\sigma|s|b, \beta, P\cup\{s\rightarrow b\})$ & \qquad \qquad if $\nexists$ path $b \rightarrow ... \rightarrow s$ in $P$ (acyclicity)\\
\textsc{reduce} $(\sigma|s, \beta, P) \Rightarrow (\sigma, \beta, P)$ \\
\qquad \qquad if $\exists k: k \rightarrow s \in P$
\\ \cline{1-1}
\textsc{arc-hybrid}\mbox{} \\\cline{1-1}
\textsc{sh} $(\sigma, b|\beta, P) \Rightarrow (\sigma|b, \beta, P)$ \\
\textsc{la}$_{e}$ $(\sigma|s, b|\beta, P) \Rightarrow (\sigma, b|\beta, P\cup\{b\rightarrow s\})$ \\
\textsc{ra}$_{st}$ $(\sigma|s_1|s_0, \beta, P) \Rightarrow (\sigma|s_1, \beta, P\cup\{s_1\rightarrow s_0\})$ \\ \bottomrule
\end{tabular}
\end{adjustbox}}
\caption{Transitions in arc-standard, arc-eager, arc-hybrid and Covington with their initial and final configurations.} 
\label{tab:transitions}
\end{table}

\subsection{Examples} 

The arc-standard transition system for projective dependency parsing (shown in Table \ref{tab:transitions}) is a left-to-right transition system, where SH is the only read transition. Figure \ref{fig:encodings} shows how a parse for an example projective sentence is converted into $n$ labels. In this transition system, $k=0$ because it needs to place words on the stack (via a read transition) before creating dependencies between them, so after having executed $i$ read transitions, the words available for parsing are restricted to $w_1 \ldots w_i$. For example, one needs two SH transitions to be able to link ``Kyrie'' as dependent of ``ate''.

The arc-eager transition system for projective dependency parsing (also in Table \ref{tab:transitions}) is a left-to-right transition system, where both SH and RA are read transitions. This shows that, while in most transition systems the notion of a read transition is implemented by a SH or shift transition (which moves a word from the remaining input into a stack or another data structure for words being processed), this need not always be the case: for example, in arc-eager, the right-arc transition also moves a word from the remaining input into the stack apart from creating a right arc. Figure \ref{fig:encodings} shows the $n$ labels resulting from the parse for the example sentence. In this transition system, $k=1$ because it can create dependencies involving the first word in the buffer, so computations with $i$ read transitions can result into a partial parse involving $w_1 \ldots w_{i+1}$. For example, the link between ``Kyrie'' and ``ate'' in Figure \ref{fig:encodings} is created after only one read transition.

\subsection{Discussion}
Informally speaking, the first condition of a left-to-right transition system simply states that when parsing a sentence of length $n$, the parser should start with a read transition and execute exactly $n$ read transitions in total. This does not impose any limits on the nature of the transitions but only on their number and arrangement in transition sequences, and it is the basis of our mapping, which uses the $n$ read transitions to split the transition sequence into $n$ subsequences that can act as labels.

The second condition means that the algorithm needs to proceed in an incremental left-to-right manner, driven by the read transitions, in the sense that each read transition introduces the possibility of using a new word in the parse, and words are introduced in left-to-right order. This generalizes the classic notion of shift transitions reading words and placing them into a data structure (like a stack) that then allows their manipulation. Note that this is a weak form of incrementality. For example, in an arc-standard dependency parser a chain of right arcs is processed from right to left, but this condition is still met.
The constant $k$ is typically $0$ or $1$ in most parsers in the literature. 
For example, it is $1$ in the classic arc-eager dependency parser and $0$ in arc-standard, as explained above. 

Moreover, we wish to stress that the notion of a left-to right transition system defined in this work only implies incrementality at the transition system level. This means that it does not guarantee that any implementation of a such transition system is left-to-right incremental in the traditional psycholinguistic sense, as the definition of a transition system only concerns the transitions available and cannot impose restrictions on what a particular implementation can use to decide between them. For instance, implementations that use BiLSTMs (as those in our own experiments below) are not left-to-right incremental, as they have information about future input encoded in the hidden representations of the input tokens. 
However, in that case, the absence of left-to-right incrementality is due to the word representations chosen for the implementation (and not due to the transition system); and any transition system that meets our definition can be implemented in a truly incremental way by using word representations that do not depend on future input.

Strictly speaking, the first condition in our definition of a left-to-right transition system is enough to define a mapping from transition sequences to sequences of $n$ labels, thus obtaining a sequence labeling encoding.
The definition of the mapping above does not formally use or rely on the second condition.
However, we include said condition because it means that the label for each word will encode information about the transitions executed after reading that word. In other words, it means that not only the transition sequences of the parser can be encoded as sequences of $n$ labels, but also that the $i$th label in the encoding is semantically linked to the $i$th word.
We believe that this is a reasonable common-sense assumption for the resulting sequence labeling model to be learnable.

An example of a transition system that fails the second condition is the easy-first parser of \newcite{goldberg-elhadad-2010-efficient}. The algorithm runs $n$ transitions in total that attach each input word
to a head, so if we ignore the second condition, all its transitions could be considered read transitions, and we would obtain a compact sequence labeling encoding where each label contains a single transition. 
However, the problem is that the label for a given word is not semantically linked to that word, as the parser can create arcs in any order, so the encoding can hardly be considered practical.

\subsection{Coverage}

\begin{table}[htbp!]
\begin{centering}
\begin{small}
\tabcolsep=0.07cm
%\resizebox{\linewidth}{!}{%
\begin{tabular}{p{50mm}lllp{50mm}lll}
\toprule
Algorithm                                & L2R? & Read t. & k & Algorithm                                & L2R? & Read t. & k  \\ \midrule

\rowcolor{Gray}
Arc-standard \cite{fraser1989parsing,nivre-2004-incrementality} & Yes            & SH               & 0  & Spinal arc-eager \cite{ballesteros-carreras-2015-transition} & Yes & SH, RA & 1 \\

Arc-eager \cite{nivre-2003-efficient} & Yes            & SH, RA           & 1  & \newcite{yamada-matsumoto-2003-statistical} & No & & \\
\rowcolor{Gray}
Arc-hybrid \cite{kuhlmann-etal-2011-dynamic} & Yes            & SH               & 1 & \newcite{choi-palmer-2011-getting} & Yes & SH & 1 \\

Covington projective \cite{covington01fundamental,nivre-2008-algorithms} & Yes & SH & 0 & \newcite{choi-mccallum-2013-transition} & Yes & No-SH, R-SH & 1 \\
\rowcolor{Gray}
Covington non-projective \cite{covington01fundamental,nivre-2008-algorithms} & Yes & SH & 0 & Non-monotonic arc-eager \cite{honnibal-etal-2013-non} & Yes & SH, RA & 1 \\

Easy-first \cite{goldberg-elhadad-2010-efficient} & No & \phantom{x} & \phantom{1} & Improved non-monotonic arc-eager \cite{honnibal-johnson-2015-improved} & No & & \\
\rowcolor{Gray}
Attardi \cite{attardi-2006-experiments} & Yes & SH & 0 & Non-monotonic Covington \cite{fernandez-gonzalez-gomez-rodriguez-2017-full} & Yes & SH & 1 \\

Planar \cite{gomez-rodriguez-nivre-2010-transition} & Yes & SH & 1 & Tree-constrained arc-eager \cite{nivre-fernandez-gonzalez-2014-squibs} & No & & \\
\rowcolor{Gray}
2-Planar \cite{gomez-rodriguez-nivre-2010-transition} & Yes & SH & 1 & Non-local Covington \cite{fernandez-gonzalez-gomez-rodriguez-2018-non} & Yes & SH & 1 \\

Arc-eager with buffer transitions \cite{fernandez-gonzalez-gomez-rodriguez-2012-improving} & Yes & SH & 2,3 & Two-register \cite{pitler-mcdonald-2015-linear} & No & & \\
\rowcolor{Gray}
Swap \cite{nivre-2009-non} & No & & & Stack-pointer \cite{ma-etal-2018-stack} & No & & \\

Swap-hybrid \cite{de-lhoneux-etal-2017-arc} & No & & & Left-to-right pointer network \cite{fernandez-gonzalez-gomez-rodriguez-2019-left} & No & & \\
\rowcolor{Gray}
Arc-swift \cite{qi-manning-2017-arc} & Yes & SH, RA$_k$ & 1 &  & & & \\

\bottomrule
\end{tabular}%
%}
\caption{Transition-based dependency parsers, whether they are left-to-right (L2R?) or not, read transitions in case they are, and value of the constant $k$. The value of $k$ should be seen as a guide only, as $k$ can vary between
variants of each parser. For example most definitions of arc-standard create arcs between nodes on the stack, so $k=0$ \cite{nivre-2004-incrementality} but it has also been defined in an equivalent form where arcs are created between stack and buffer, so $k=1$ (see \cite{nivre-2008-algorithms}). The same happens with Attardi and other algorithms.}
\label{table:depparsers}
\end{small}
\end{centering}
\end{table}

Table \ref{table:depparsers} shows different transition systems from the dependency parsing literature and whether they are left-to-right or not, together with the values of the read transitions and $k$ if applicable. As can be seen in the table, the majority of known transition-based parsers conform to our definition of left-to-right, and hence yield an encoding that can be used to define a sequence labeling parser with the framework defined here. Exceptions are parsers with multiple left-to-right passes over the input \cite{yamada-matsumoto-2003-statistical}, those that can create arcs between words in arbitrary order (like the aforementioned easy-first parser of \newcite{goldberg-elhadad-2010-efficient}) those that use unshift transitions that return a node to the buffer \cite{nivre-fernandez-gonzalez-2014-squibs,honnibal-johnson-2015-improved} or swap transitions which also have that side effect \cite{nivre-2009-non}, and those that can create arcs involving nodes arbitrarily far to the right \cite{ma-etal-2018-stack,fernandez-gonzalez-gomez-rodriguez-2019-left}. 

For all of these exceptions, one could still construct sequence-labeling encodings by ignoring the second condition of a left-to-right system as arc-creating transitions always meet the first condition, but this is dubiously practical in most cases, as explained above. A singular case in this respect is the left-to-right pointer network parser of \newcite{fernandez-gonzalez-gomez-rodriguez-2019-left}. This parser does not fit the second condition of our definition because it can use nodes arbitrarily to the right as heads (in spite of being purely left-to-right in terms of the order in which it considers dependents); and if we still apply our transformation to it, we obtain an encoding isomorphic to the relative positional encoding used in \newcite{li-etal-2018-seq2seq}, which has been shown useful under strong machine learning models \cite{vacareanu-etal-2020-parsing}.

\begin{table}[tb]
\begin{scriptsize}
\begin{centering}
%\resizebox{\columnwidth}{!}{%
\begin{tabular}{p{50mm}lllp{50mm}lll}
\toprule
Algorithm                                & L2R? & R. t. & k & Algorithm                                & L2R? & R. t. & k  \\ \midrule
\rowcolor{Gray}
Bottom-up \cite{sagae-lavie-2005-classifier} & Yes & SH & 0 &In-order \cite{liu-zhang-2017-order} & Yes & SH & 0 \\

Bottom-up with terminate action \cite{zhang-clark-2009-transition} & Yes & SH & 0 & Discontinuous easy-first \cite{versley-2014-experiments} & No & & \\
\rowcolor{Gray}
Bottom-up with padding \cite{zhu-etal-2013-fast} & Yes & SH & 0 & Swap \cite{maier-2015-discontinuous} & No & & \\

Odd-even \cite{mi-huang-2015-shift} & Yes & SH & 0 & Skip-shift \cite{maier-lichte-2016-discontinuous} & No & & \\
\rowcolor{Gray}
LR-inspired \cite{crabbe-2014-lr} & Yes & SH & 0 & \newcite{stanojevic-g-alhama-2017-neural} & No & & \\

Span-based \cite{cross-huang-2016-span} & Yes & SH & 0 & SR-GAP \cite{coavoux-crabbe-2017-incremental} & Yes & SH & 0 \\
\rowcolor{Gray}
\newcite{coavoux-crabbe-2016-neural} & Yes & SH & 0 & ML-GAP and ML-GAP-LEX \cite{coavoux-etal-2019-unlexicalized} & Yes & SH & 0 \\

Non-binary bottom-up \cite{FerGomAI2019} & Yes & SH & 0 & \newcite{coavoux-cohen-2019-discontinuous} stack-free system & Yes & SH & 0 \\
\rowcolor{Gray}
Top-down \cite{dyer-etal-2016-recurrent} & Yes & SH & 0 & Tetra-tagging \cite{tetra} & Yes & {\tiny$\nwarrow$}, {\tiny$\nearrow$} & 0 \\

\bottomrule
\end{tabular}%
%}
\end{centering}
\caption{Transition-based constituent parsers, whether they are left-to-right (L2R?) or not, read transitions in case they are, and value of the constant $k$.}
\label{table:constparsers}
\end{scriptsize}
\end{table}

Table \ref{table:constparsers} shows transition-based constituency parsers %in the literature, 
with analogous information to Table \ref{table:depparsers}. 
In this case, 
all of the listed parsers that do not fall into our definition of left-to-right are discontinuous constituent parsers that use swap transitions \cite{versley-2014-experiments,maier-2015-discontinuous}. Every continuous constituent parser that we found is covered, as well as discontinuous parsers that use other devices, like gap transitions \cite{coavoux-etal-2019-unlexicalized} or set handling \cite{coavoux-cohen-2019-discontinuous}. All of the supported constituent parsers have $k=0$, following the traditional shift-reduce paradigm that only operates on nodes after reading them from the input buffer.

\begin{table}[htpb]
\begin{small}
%\resizebox{\columnwidth}{!}{%
\begin{tabular}{p{37mm}lllp{45mm}lll}
\toprule
Algorithm                                & L2R? & Read t. & k & Algorithm                                & L2R? & Read t. & k  \\ \midrule
\rowcolor{Gray}
\newcite{sagae-tsujii-2008-shift} & Yes & SH & 1 & Online reordering \cite{zhang-etal-2016-transition} & No & & \\

\newcite{titov2009online} & Yes & SH & 1 & Two-stack \cite{zhang-etal-2016-transition} & Yes & SH & 1 \\
\rowcolor{Gray}
\newcite{ribeyre-etal-2014-alpage}&No&&& \newcite{hershcovich-etal-2017-transition}&No&&\\

\newcite{tokgoz-eryigit-2015-transition} & Yes & SH & 1 & \newcite{Wang2018} & Yes & No-SH, R-SH & 1 \\
\rowcolor{Gray}
\newcite{artsymenia-etal-2016-ihs}&Yes&SH&1& DM, PSD \cite{che-etal-2019-hit} & Yes & SH & 1\\
\bottomrule
\end{tabular}%
%}
\caption{Transition-based semantic dependency parsers, whether they are left-to-right (L2R?) or not, read transitions in case they are, and value of the constant $k$. For simplicity we only include semantic dependency parsers and exclude parsers for formalisms that go beyond dependency graphs, like AMR, and cross-framework parsers 
(e.g. \newcite{bai-zhao-2019-sjtu}).
}
\label{table:semparsers}
\end{small}
\end{table}

Table \ref{table:semparsers} lists transition-based semantic dependency parsers with analogous information to Tables \ref{table:depparsers} and \ref{table:constparsers}. 
Once again,
parsers that do not fall into our definition of left-to-right are mostly those that define a swap transition. Note that for this coverage analysis we exclude semantic formalisms that go beyond dependency graphs, such as AMR \cite{banarescu-etal-2013-abstract}, where pure transition-based models (e.g. \newcite{damonte-etal-2017-incremental}, \newcite{ballesteros-al-onaizan-2017-amr},\newcite{vilares-gomez-rodriguez-2018-transition}) need to remove tokens and also create (multiple) concepts from words, and therefore include specific transitions to do so. Removing tokens can be seen as a read transition, but creating many concept nodes from a single word
breaks the left-to-right condition and 
does not ensure that the system will have $n$ read transitions, where $n$ is the length of the input sentence. Although outside the scope of this paper, note that a hybrid system that first computed the concepts from words (e.g. with a seq2seq architecture), to then apply a transition-based graph parser could support the left-to-right condition in the same way as other formalisms we have considered.

Putting it all together, our mapping is applicable to a wide range of transition systems, spanning a variety of formalisms: projective and non-projective dependency parsing, continuous and discontinuous constituency parsing, and various flavours of semantic parsing.
Also, note that the impossibility to map buffer-based\footnote{With buffer-based we refer to a swap transition that puts back a word from the stack to the buffer, contrary to stack-based swap that changes the order of the words in the stack, respecting the left-to-right condition. 
} swap transition-based parsers is not related to inability of our approach to handle non-projectivity,
but to the non-left-to-right nature of those swap models. For instance, in Table \ref{table:depparsers} we showed how other non-projective transition-based algorithms \cite{covington01fundamental,attardi-2006-experiments,gomez-rodriguez-nivre-2010-transition} can be mapped to a sequence labeling encoding.

\section{Experiments}

To test the practical applicability of our theoretical contribution, we implement sequence labeling versions, obtained using the mapping in Section \ref{sec:mapping}, of various syntactic dependency parsers. We include three well-known projective parsers: arc-standard \cite{nivre-2004-incrementality}, arc-eager \cite{nivre-2003-efficient} and arc-hybrid \cite{kuhlmann-etal-2011-dynamic}; and one parser with full coverage of non-projective trees: the Covington non-projective parser \cite{covington01fundamental,nivre-2008-algorithms}. The transition systems for all of these parsers are shown in Table \ref{tab:transitions}.
For comparison, we also include two of the best encodings to the date for dependency parsing as labeling \cite{strzyz-etal-2019-viable}: (i) the rel-PoS \cite{spoustova2010dependency} and (ii) the bracketing encoding \cite{yli-jyra-gomez-rodriguez-2017-generic}. The former casts the problem as a head-selection task where each token encodes its head using a PoS-tag-based offset. 
The latter assigns a label to each token encoding a set of incoming/outgoing arcs from that and neighbouring tokens.

\subsection{Data}

Following \newcite{anderson2020distilling}, we choose a subset of UDv2.4 treebanks \cite{ud2.4} which includes languages with a variety of corpus sizes, language typologies, alphabets and levels of non-projectivity, among other differences. More particularly,
these treebanks are: Ancient Greek\textsubscript{Perseus}, Chinese\textsubscript{GSD}, English\textsubscript{EWT}, Finnish\textsubscript{TDT}, Hebrew\textsubscript{HTB}, Russian\textsubscript{GSD}, Tamil\textsubscript{TTB}, Uyghur\textsubscript{UDT} and Wolof\textsubscript{WTB}. Appendix \ref{appendix-label-sizes} shows the number of labels that our approach generates for each treebank.
We also use UDpipe \cite{straka-2018-udpipe} to obtain data with predicted segmentation, tokenization and PoS tags.

\newcolumntype{g}{>{\columncolor{Gray}}c}
\begin{table*}[hptb!]
\centering
\small{
\tabcolsep=0.065cm
\begin{tabular}{ll|ggccggccggccggccggcc}
\toprule
\multirow{3}{*}{Split}&\multirow{3}{*}{Encoding}&\multicolumn{2}{g}{Ancient }&\multicolumn{2}{c}{Chinese}&\multicolumn{2}{g}{English}&\multicolumn{2}{c}{Finnish}&\multicolumn{2}{g}{Hebrew}&\multicolumn{2}{c}{Russian}&\multicolumn{2}{g}{Tamil}&\multicolumn{2}{c}{Uyghur}&\multicolumn{2}{g}{Wolof}&\multicolumn{2}{c}{Avg}\\
&&\multicolumn{2}{g}{Greek}&&&&&&&&&&&&&&&&\\

&&\tiny{UAS}&\tiny{LAS}&\tiny{UAS}&\tiny{LAS}&\tiny{UAS}&\tiny{LAS}&\tiny{UAS}&\tiny{LAS}&\tiny{UAS}&\tiny{LAS}&\tiny{UAS}&\tiny{LAS}&\tiny{UAS}&\tiny{LAS}&\tiny{UAS}&\tiny{LAS}&\tiny{UAS}&\tiny{LAS}&\tiny{UAS}&\tiny{LAS}\\
\hline
\multirow{6}{*}{dev\textsubscript{s\ref{bilstm-setup-1}}}&rel-PoS & 70.2&	61.8&	61.6&	57.4&	82.6&	79.0&	83.2&	79.3&	67.4&	63.8&	84.0&	79.6&	59.5&	52.0&	72.6&	59.0&	77.0&	70.5& \textit{73.1} & \textit{66.9}\\
&Brackets & 65.2&	57.2&	61.8&	57.9&	82.2&	78.6&	82.8&	79.1&	67.1&	63.4&	83.5&	78.9&	57.4&	50.2&	73.1&	59.9&	76.3&	69.7& \textit{72.2}&\textit{66.1}\\
&Standard &62.2&	54.7&	60.3&	56.2&	79.9&	76.3&	79.7&	76.0&	64.2&	60.5&	80.0&	75.8&	57.8&	49.7&	72.5&	59.2&	73.0&	66.8&	\textit{69.9}&	\textit{63.9}\\
&Eager &60.4&	52.9&	60.9&	56.9&	80.8&	77.3&	80.2&	76.4&	65.4&	61.8&	81.5&	77.4&	57.9&	50.2&	73.2&	59.9&	74.6&	68.0&	\textit{70.5}&	\textit{64.5}\\
&Hybrid &62.7&	55.2&	60.3&	56.3&	80.5&	76.9&	80.4&	76.7&	65.2&	61.6&	81.1&	76.4&	58.1&	50.5&	72.9&	59.5&	72.9&	66.5&	\textit{70.5}&	\textit{64.4}\\ 
&Covington &64.0&	56.0&	56.1&	52.3&	79.6&	75.9&	80.7&	76.6&	64.7&	61.1&	80.0&	75.7&	53.2&	45.7&	67.2&	55.0&	71.7&	65.3&\textit{68.6}& \textit{62.6} \\
\hline
\multirow{6}{*}{test\textsubscript{s\ref{bilstm-setup-1}}}&rel-PoS & 69.5&	60.0&	64.4&	60.0&	82.0&	78.5&	80.2&	74.9&	63.0&	59.5&	82.7&	77.3&	58.0&	49.8&	71.5&	57.8&	76.1&	69.6 &\textit{71.9}&\textit{65.3}\\
&Brackets &64.9&	56.2&	64.2&	59.8&	81.7&	78.3&	81.1&	76.0&	62.1&	58.7&	82.9&	77.9&	57.2&	49.0&	72.0&	58.5&	75.4&	68.8&\textit{71.3}&\textit{64.8} \\
&Standard & 61.0&	52.6&	62.9&	58.5&	79.2&	75.8&	78.5&	73.4&	60.3&	56.5&	79.1&	74.2&	57.1&	48.6&	71.5&	58.2&	72.0&	65.7&	\textit{69.1}&	\textit{62.6}\\
&Eager & 60.4&	52.0&	63.2&	59.0&	80.2&	76.7&	79.0&	73.6&	61.2&	57.6&	80.4&	75.3&	56.2&	48.3&	71.9&	58.7&	73.5&	67.1&	\textit{69.6}&	\textit{63.1}\\
&Hybrid & 62.1&	53.5&	63.2&	58.2&	79.5&	76.1&	78.9&	74.0&	60.4&	56.8&	79.3&	74.4&	57.0&	48.7&	71.3&	58.6&	72.2&	65.7&	\textit{69.3}&	\textit{62.9}\\
&Covington &63.7&	54.4&	58.3&	54.2&	78.5&	75.0&	78.9&	73.7&	60.2&	56.6&	79.2&	74.0&	52.3&	44.8&	66.4&	53.5&	69.9&	63.8&\textit{67.5}&\textit{61.1} \\
\hline
\hline
\multirow{6}{*}{dev\textsubscript{s\ref{bilstm-setup-2}}}&rel-PoS &65.3&	58.3&	58.8&	55.3&	80.3&	77.1&	80.8&	77.3&	65.3&	62.2&	81.3&	77.4&	52.7&	41.9&	65.7&	53.3&	73.1&	66.3& \textit{69.2}&\textit{63.2}\\
&Brackets & 64.7&	57.2&	63.8&	59.4&	83.4&	80.0&	84.1&	80.4&	68.7&	65.0&	84.0&	79.7&	55.9&	45.1&	72.1&	58.8&	75.2&	67.4&\textit{72.4}&\textit{65.9}\\
&Standard & 61.6&	54.8&	61.9&	57.6&	80.9&	77.5&	80.7&	77.2&	65.9&	62.6&	80.3&	76.2&	55.0&	43.3&	71.6&	58.2&	70.4&	63.2& \textit{69.8}&\textit{63.4} \\
&Eager & 60.0&	53.2&	63.4&	59.2&	81.9&	78.5&	81.6&	78.2&	66.9&	63.2&	81.7&	77.4&	56.4&	45.5&	72.3&	58.8&	73.4&	66.1&	\textit{70.9}&	\textit{64.4}\\
&Hybrid & 62.3&	55.5&	62.4&	58.1&	81.4&	78.0&	81.3&	77.9&	66.1&	62.6&	81.0&	76.9&	56.3&	44.9&	71.7&	58.1&	71.9&	64.8&	\textit{70.5}&	\textit{64.1}\\
&Covington &63.7&	56.1&	57.4&	53.5&	80.5&	76.8&	81.8&	78.0&	66.3&	63.0&	80.0&	75.7&	51.1&	40.8&	67.0&	54.1&	71.3&	63.6&	\textit{68.8}&	\textit{62.4} \\
\hline
\multirow{6}{*}{test\textsubscript{s\ref{bilstm-setup-2}}}&rel-PoS & 62.9& 	55.1& 	60.3& 	56.8& 	78.5& 	75.3& 	65.6& 	59.6& 	59.7& 	56.6& 	79.0& 	73.8& 	51.6& 	40.6& 	64.9& 	52.4& 	72.3& 	65.6& 	\textit{66.1}& 	\textit{59.5}\\
&Brackets & 63.4&	54.8&	65.3&	61.1&	82.4&	78.9&	75.1&	67.7&	62.3&	58.6&	81.8&	75.8&	56.9&	42.6&	71.0&	57.4&	75.1&	67.2&	\textit{70.4}&	\textit{62.7} \\
&Standard & 59.5&	52.0&	64.2&	59.9&	79.3&	75.8&	73.4&	66.4&	60.1&	56.4&	78.6&	73.1&	53.8&	41.3&	70.5&	56.9&	71.1&	63.6&	\textit{67.8}&	\textit{60.6} \\
&Eager & 59.5&	51.7&	64.5&	60.5&	80.0&	76.6&	72.3&	64.1&	61.1&	57.3&	79.7&	74.0&	53.4&	41.1&	71.6&	57.9&	72.9&	65.2&	\textit{68.3}&	\textit{60.9}\\
&Hybrid & 60.4&	52.5&	64.2&	59.9&	80.1&	76.7&	74.7&	67.5&	60.6&	57.0&	78.6&	73.1&	53.7&	41.7&	71.1&	57.6&	71.5&	63.7&	\textit{68.3}&	\textit{61.1} \\
&Covington &62.5&	53.5&	59.6&	55.7&	78.6&	74.9&	73.8&	66.6&	60.6&	57.0&	78.2&	72.5&	48.2&	36.6&	66.3&	53.6&	69.2&	61.6&	\textit{66.3}&	\textit{59.1} \\
\bottomrule

\end{tabular}}
\caption{Results for the BiLSTM setups \ref{bilstm-setup-1} (s1, with PoS tags) and \ref{bilstm-setup-2} (s2, without PoS tags).}
\label{tab:main-results-ncrf}
\end{table*}

\subsection{Sequence labeling models}\label{section-sequence-labeling-models}

For training, we consider two sequence labeling encoders (see Appendix \ref{appendix-hyperparameters} for hyper-parameters), which will produce $n$ hidden contextualized representations $\vec{h}_i$ to generate the labels: 

\paragraph{BiLSTMs}  \cite{hochreiter1997long,schuster1997bidirectional} We use the NCRF++ framework \cite{yang-zhang-2018-ncrf} to train the models and consider two different setups: 
\begin{enumerate}
    \item\label{bilstm-setup-1} A setup where the input vectors to the models are a concatenation of the pre-trained word embeddings by \newcite{ginter-et-al-2017-conll}, 
    a second word vector generated by a char-LSTM layer, which is trained together with the rest of the network, and PoS tag vectors.
    \item\label{bilstm-setup-2} Same as \ref{bilstm-setup-1}, but without PoS tag vectors.
\end{enumerate}

The motivation for setup \ref{bilstm-setup-1} is that the PoS-tag-based encoding \cite{strzyz-etal-2019-viable} requires PoS tags to decode the labels into a tree. Therefore, as PoS tags need to be computed to decode the tree, it is a fair argument to use them as input parameters, as done by \newcite{strzyz-etal-2019-viable}.
 The motivation for setup \ref{bilstm-setup-2} is that in an era where the usefulness of PoS tags has been 
 questioned \cite{de-lhoneux-etal-2017-raw,anderson2020frailty}, our encodings (which do not require PoS tags to decode the trees) could benefit in terms of speed and simplicity, with a minimum cost to accuracy.

\paragraph{BERT} \cite{devlin-etal-2019-bert} By default we fine-tune multi-lingual BERT (M-BERT) and in particular bert-base-multilingual-cased, 
except for English, Chinese and Finnish \cite{virtanen2019multilingual}; for which monolingual models are available.\footnote{\url{https://huggingface.co/transformers/pretrained_models.html}.}\textsuperscript{,}\footnote{bert-base-cased, bert-base-chinese, and TurkuNLP/bert-base-finnish-cased-v1}
BERT splits the input into sub-word pieces \cite{wu2016google} generating more sub-words than tokens, while we have a fixed amount of labels (equal to the number of tokens in the sentence) to assign. To solve this, we consider a word to be its first sub-word.
Contrary to BiLSTMs, BERT does not use PoS tags as input, and we will only use them to decode the trees of the rel-PoS encoding.\\

\noindent To generate the output labels from the hidden vectors, we map each $\vec{h_i}$ to two output softmax layers (tasks) following a hard-sharing multi-task learning architecture: one that predicts the subsequence of transitions associated to that word, and a second one that predicts the dependency relation between that word and its head. The loss is computed as the sum of the categorical cross-entropy of both tasks. Corrupted sequences of predicted labels are postprocessed according to Appendix \ref{appendix-postprocessing}.

\begin{table*}[hbtp!]
\centering
\small{
\renewcommand{\arraystretch}{0.9}
\tabcolsep=0.15cm
\begin{tabular}{ll|ggccggccggccgg}
%&\multicolumn{12}{c}{Treebank}\\
\toprule
\multirow{2}{*}{Split}&\multirow{2}{*}{Encoding}&\multicolumn{2}{g}{Chinese}&\multicolumn{2}{c}{English}&\multicolumn{2}{g}{Finnish}&\multicolumn{2}{c}{Hebrew}&\multicolumn{2}{g}{Russian}&\multicolumn{2}{c}{Tamil}&\multicolumn{2}{g}{Avg}\\
&&\tiny{UAS}&\tiny{LAS}&\tiny{UAS}&\tiny{LAS}&\tiny{UAS}&\tiny{LAS}&\tiny{UAS}&\tiny{LAS}&\tiny{UAS}&\tiny{LAS}&\tiny{UAS}&\tiny{LAS}&\tiny{UAS}&\tiny{LAS}\\
\hline
\multirow{6}{*}{dev}&rel-PoS &59.8&	57.0&	82.3&	79.9&	82.6&	79.7&	61.8&	58.0&	77.6&	73.2&	43.4&	33.6&	\textit{67.9}&	\textit{63.5}\\
&Brackets &69.3&	65.3&	87.2&	84.4&	88.6&	85.1&	66.6&	61.7&	80.6&	75.4&	49.8&	36.8&	\textit{73.7}&	\textit{68.1}\\
&Standard &67.9&	64.1&	85.4&	82.6&	85.7&	82.5&	64.4&	59.8&	77.8&	72.7&	48.2&	35.1&	\textit{71.6}&	\textit{66.1} \\
&Eager &67.0&	63.2&	85.3&	82.4&	86.8&	83.5&	64.7&	59.6&	77.8&	72.9&	49.7&	35.2&	\textit{71.9}&	\textit{66.1} \\
&Hybrid &67.0&	63.3&	85.5&	82.8&	86.6&	83.1&	65.2&	60.1&	78.1&	73.4&	47.3&	34.0&	\textit{71.6}&	\textit{66.1} \\
&Covington &59.7&	56.0&	81.4&	78.8&	82.4&	79.2&	61.9&	57.3&	73.9&	68.9&	44.6&	31.6&	\textit{67.3}&	\textit{62.0} \\
\hline
\hline
\multirow{6}{*}{test}&rel-PoS &61.5&	58.8&	81.0&	78.8&	82.6&	80.0&	56.9&	53.1&	77.9&	73.2&	45.2&	34.1&	\textit{67.5}&	\textit{63.0}\\
&Brackets &70.2&	66.5&	86.2&	83.3&	89.4&	86.1&	62.7&	57.9&	79.9&	74.2&	48.9&	35.9&	\textit{72.9}&	\textit{67.3} \\
&Standard &68.5 &	64.7 &	84.0 &	81.2 &	86.1 &	83.1 &	59.6 &	54.3 &	77.6 &	72.3 &	51 &	37.3 &	\textit{71.1} &	\textit{65.5}\\
&Eager &69.2&	65.6&	83.8&	81.2&	87.2&	83.9&	60.2&	55.3&	77.6&	72.4&	48.6&	36.7&	\textit{71.1}&	\textit{65.8}\\
&Hybrid &69.1&	65.5&	84.1&	81.3&	86.6&	83.6&	60.3&	55.7&	77.5&	72.3&	48.2&	35.6&	\textit{71.0}&	\textit{65.7}\\
&Covington &60.1&	56.7&	79.7&	77.4&	83.2&	80.2&	57.6&	52.6&	73.1&	68.0&	46.3&	34.0&	\textit{66.7}&	\textit{61.5} \\
\bottomrule
\end{tabular}}
\caption{Results for the BERT setup.}
\label{tab:main-results-bert}
\end{table*}

\subsection{Results}

Table \ref{tab:main-results-ncrf} shows the results of the encodings trained with the BiLSTM architecture. Overall, we see that the transition-based encodings perform comparable to preexisting encodings. The performance of arc-standard, arc-eager and arc-hybrid is similar across the board, while Covington suffers more, probably due to the large output vocabulary (see Appendix \ref{appendix-label-sizes}) as a Covington transition sequence has $O(n^2)$ transitions that we are grouping into $n$ labels, contrary to the other three algorithms which run $O(n)$ transitions per sentence.
Still, for highly non-projective treebanks such as Ancient-Greek-Perseus, the Covington mapping performs the best among the transition-based encodings; showing that it is learnable and yet preferable in this case to using pure projective transition-based encodings.

With respect to the BiLSTM setups \ref{bilstm-setup-1} (with PoS tags) and \ref{bilstm-setup-2} (without PoS tags), we observe that while the transition-based encodings (and the bracketing-based one) suffer little from the absence of PoS-tags, the rel-PoS based one greatly needs them, losing over 5 LAS points on average when they are not used. This is relevant since the use of PoS-tags increases the latency, and their usefulness has been questioned in some parsing setups \cite{de-lhoneux-etal-2017-raw,smith-etal-2018-investigation}. In addition, in Appendix \ref{upperbounds} we 
compute empirical upper bounds for the accuracy of the encodings by running them on the gold datasets (gold segmentation, tokenization and PoS tags) and compare them with the accuracy of the predicted setup with PoS tags. We can observe that the accuracy of upstream tasks has a large impact on all models, including the baselines, with the largest gaps between the parsing results on the predicted and gold dataset being observed for treebanks where segmentation, tokenization or tagging is difficult (for instance, the biggest difference is observed in Hebrew, where words and UPoS have the lowest prediction accuracy).

Table \ref{tab:main-results-bert} shows the results for BERT. We excluded Ancient-Greek, Uyghur and Wolof since M-BERT does not support them, and we are not aware of a monolingual model. The tendency is similar to the BiLSTM setup \ref{bilstm-setup-2}, but with higher scores for monolingual BERTs (not M-BERT though) and rel-PoS lagging further behind the bracketing and the projective transition-based encodings (on average around 4 and 2.5 LAS points, respectively).

It is also worth comparing the results of our sequence labeling implementations of the projective transition-based parsers to the experiments of \newcite{shi-etal-2017-fast} with regular transition-based implementations and different sets of positional features. For this, we run our BiLSTM setup on the same dataset as them, i.e., the English PTB dev set. The results can be seen in Table \ref{tab:shi}. \newcite{shi-etal-2017-fast} concluded that, with a BiLSTM-based architecture, two positional features (one stack and one buffer feature) were needed to obtain reasonable accuracy in the arc-eager and arc-hybrid parsers, and three (two stack and one buffer feature) in the case of the arc-standard parser. In contrast, and although the exact accuracy numbers do not provide a homogeneous comparison due to hyperparameter differences,
it can be seen in Table \ref{tab:shi} that our labels of sequences of transitions can be learned with only one positional feature (our setup just assigns a sequence of labels beginning with a SH to each word shifted from the input, i.e., it only uses the first 
buffer word $b_0$, and has access to no explicit representation of stack elements at all). Thus, this shows
that our multi-transition labels seem to be learnable with less data than individual transitions in transition systems, although the latter (with suitable features) are still ahead in terms of raw accuracy.
While similar effects had been observed in seq2seq transition-based parsers \cite{zhang-etal-2017-stack,liu-zhang-2017-encoder}, these use more complex neural architectures than transition-based implementations, including attention weights that can focus on words in prominent stack positions \cite{liu-zhang-2017-encoder}. Here, we are just using plain BiLSTMs as in the standard transition-based implementation of \newcite{shi-etal-2017-fast}.

\begin{table}[hbtp]
\centering
\small{
\begin{tabular}{@{}lrccc@{}}
\toprule
Model & \multicolumn{1}{c}{Features} & Arc-standard & Arc-eager & Arc-hybrid \\ \midrule
\multirow{4}{*}{Shi et al.} & \{$s_2$, $s_1$, $s_0$, $b_0$\} & \stddev{93.95}{0.12} & \stddev{93.92}{0.04} & \stddev{94.08}{0.13} \\
 & \{$s_1$, $s_0$, $b_0$\} & \stddev{94.13}{0.06} & \stddev{93.91}{0.07} & \stddev{94.08}{0.05} \\
 & \{$s_0$, $b_0$\} & \stddev{54.47}{0.36} & \stddev{93.92}{0.07} & \stddev{94.03}{0.12} \\
 & \{$b_0$\} & \stddev{47.11}{0.44} & \stddev{79.15}{0.06} & \stddev{52.39}{0.23} \\ \midrule
Our model & \{$b_0$\} & 92.13 & 93.22 &92.66  \\ \bottomrule
\end{tabular}}
\caption{Performance (in UAS\%) of our system using a single positional feature $b_0$ compared with \newcite{shi-etal-2017-fast} on the English PTB dev set. }
\label{tab:shi}
\end{table}

\section{Conclusion}

This paper has established a theoretical relationship between transition-based and sequence labeling parsing valid for a broad definition of left-to-right transition-based algorithms. It also provides a new set of encodings for sequence labeling parsing which are automatically derived, in contrast to existing ones which were created ad-hoc for this purpose. To test the practical utility, we ran experiments on dependency parsing for a diverse set of languages. 
Interestingly, the mapping is meaningful and learnable, despite not using any representation of stack nodes.
While our experiments focused on dependency parsing as we only aimed to validate the theory, an obvious avenue for future work is to implement and test the sequence-labeling encodings derived from the constituent and semantic parsers in Tables \ref{table:constparsers} and \ref{table:semparsers}.
The latter are particularly relevant in the sense that, to our knowledge, there is no previous work on semantic parsing as sequence labeling; so our method provides the first encodings for this purpose.

\section*{Acknowledgments}

This work has received funding from the European Research Council (ERC), which has funded this research under the European Union's Horizon 2020 research and innovation programme (FASTPARSE, grant agreement No 714150), from MINECO (ANSWER-ASAP, TIN2017-85160-C2-1-R), from Xunta de Galicia (ED431C 2020/11), and from Centro de Investigación de Galicia `CITIC', funded by Xunta de Galicia and the European Union (European Regional Development Fund- Galicia 2014-2020 Program), by grant ED431G 2019/01. DV is supported by a 2020 Leonardo Grant for Researchers and Cultural Creators from the BBVA Foundation.

\bibliographystyle{coling}
\bibliography{coling2020,anthology}

\begin{thebibliography}{}

\bibitem[\protect\citename{Anderson and
  G{\'o}mez-Rodr{\'i}guez}2020]{anderson2020distilling}
Mark Anderson and Carlos G{\'o}mez-Rodr{\'i}guez.
\newblock 2020.
\newblock Distilling neural networks for greener and faster dependency parsing.
\newblock In {\em Proceedings of the 16th International Conference on Parsing
  Technologies and the IWPT 2020 Shared Task on Parsing into Enhanced Universal
  Dependencies}, pages 2--13, Online, July. Association for Computational
  Linguistics.

\bibitem[\protect\citename{Anderson and
  Gómez-Rodríguez}2020]{anderson2020frailty}
Mark Anderson and Carlos Gómez-Rodríguez.
\newblock 2020.
\newblock {On the Frailty of Universal POS Tags for Neural UD Parsers}.
\newblock In {\em Proceedings of the 24th Conference on Computational Natural
  Language Learning}.
\newblock To appear.

\bibitem[\protect\citename{Artsymenia \bgroup et al.\egroup
  }2016]{artsymenia-etal-2016-ihs}
Artsiom Artsymenia, Palina Dounar, and Maria Yermakovich.
\newblock 2016.
\newblock {IHS}-{RD}-belarus at {S}em{E}val-2016 task 9: Transition-based
  {C}hinese semantic dependency parsing with online reordering and
  bootstrapping.
\newblock In {\em Proceedings of the 10th International Workshop on Semantic
  Evaluation ({S}em{E}val-2016)}, pages 1207--1211, San Diego, California,
  June. Association for Computational Linguistics.

\bibitem[\protect\citename{Attardi}2006]{attardi-2006-experiments}
Giuseppe Attardi.
\newblock 2006.
\newblock Experiments with a multilanguage non-projective dependency parser.
\newblock In {\em Proceedings of the Tenth Conference on Computational Natural
  Language Learning ({C}o{NLL}-X)}, pages 166--170, New York City, June.
  Association for Computational Linguistics.

\bibitem[\protect\citename{Bai and Zhao}2019]{bai-zhao-2019-sjtu}
Hongxiao Bai and Hai Zhao.
\newblock 2019.
\newblock {SJTU} at {MRP} 2019: A transition-based multi-task parser for
  cross-framework meaning representation parsing.
\newblock In {\em Proceedings of the Shared Task on Cross-Framework Meaning
  Representation Parsing at the 2019 Conference on Natural Language Learning},
  pages 86--94, Hong Kong, November. Association for Computational Linguistics.

\bibitem[\protect\citename{Ballesteros and
  Al-Onaizan}2017]{ballesteros-al-onaizan-2017-amr}
Miguel Ballesteros and Yaser Al-Onaizan.
\newblock 2017.
\newblock {AMR} parsing using stack-{LSTM}s.
\newblock In {\em Proceedings of the 2017 Conference on Empirical Methods in
  Natural Language Processing}, pages 1269--1275, Copenhagen, Denmark,
  September. Association for Computational Linguistics.

\bibitem[\protect\citename{Ballesteros and
  Carreras}2015]{ballesteros-carreras-2015-transition}
Miguel Ballesteros and Xavier Carreras.
\newblock 2015.
\newblock Transition-based spinal parsing.
\newblock In {\em Proceedings of the Nineteenth Conference on Computational
  Natural Language Learning}, pages 289--299, Beijing, China, July. Association
  for Computational Linguistics.

\bibitem[\protect\citename{Banarescu \bgroup et al.\egroup
  }2013]{banarescu-etal-2013-abstract}
Laura Banarescu, Claire Bonial, Shu Cai, Madalina Georgescu, Kira Griffitt, Ulf
  Hermjakob, Kevin Knight, Philipp Koehn, Martha Palmer, and Nathan Schneider.
\newblock 2013.
\newblock Abstract meaning representation for sembanking.
\newblock In {\em Proceedings of the 7th Linguistic Annotation Workshop and
  Interoperability with Discourse}, pages 178--186, Sofia, Bulgaria, August.
  Association for Computational Linguistics.

\bibitem[\protect\citename{Che \bgroup et al.\egroup }2019]{che-etal-2019-hit}
Wanxiang Che, Longxu Dou, Yang Xu, Yuxuan Wang, Yijia Liu, and Ting Liu.
\newblock 2019.
\newblock {HIT}-{SCIR} at {MRP} 2019: A unified pipeline for meaning
  representation parsing via efficient training and effective encoding.
\newblock In {\em Proceedings of the Shared Task on Cross-Framework Meaning
  Representation Parsing at the 2019 Conference on Natural Language Learning},
  pages 76--85, Hong Kong, November. Association for Computational Linguistics.

\bibitem[\protect\citename{Choi and
  McCallum}2013]{choi-mccallum-2013-transition}
Jinho~D. Choi and Andrew McCallum.
\newblock 2013.
\newblock Transition-based dependency parsing with selectional branching.
\newblock In {\em Proceedings of the 51st Annual Meeting of the Association for
  Computational Linguistics (Volume 1: Long Papers)}, pages 1052--1062, Sofia,
  Bulgaria, August. Association for Computational Linguistics.

\bibitem[\protect\citename{Choi and Palmer}2011]{choi-palmer-2011-getting}
Jinho~D. Choi and Martha Palmer.
\newblock 2011.
\newblock Getting the most out of transition-based dependency parsing.
\newblock In {\em Proceedings of the 49th Annual Meeting of the Association for
  Computational Linguistics: Human Language Technologies}, pages 687--692,
  Portland, Oregon, USA, June. Association for Computational Linguistics.

\bibitem[\protect\citename{Coavoux and
  Cohen}2019]{coavoux-cohen-2019-discontinuous}
Maximin Coavoux and Shay~B. Cohen.
\newblock 2019.
\newblock Discontinuous constituency parsing with a stack-free transition
  system and a dynamic oracle.
\newblock In {\em Proceedings of the 2019 Conference of the North {A}merican
  Chapter of the Association for Computational Linguistics: Human Language
  Technologies, Volume 1 (Long and Short Papers)}, pages 204--217, Minneapolis,
  Minnesota, June. Association for Computational Linguistics.

\bibitem[\protect\citename{Coavoux and
  Crabb{\'e}}2016]{coavoux-crabbe-2016-neural}
Maximin Coavoux and Beno{\^\i}t Crabb{\'e}.
\newblock 2016.
\newblock Neural greedy constituent parsing with dynamic oracles.
\newblock In {\em Proceedings of the 54th Annual Meeting of the Association for
  Computational Linguistics (Volume 1: Long Papers)}, pages 172--182, Berlin,
  Germany, August. Association for Computational Linguistics.

\bibitem[\protect\citename{Coavoux and
  Crabb{\'e}}2017]{coavoux-crabbe-2017-incremental}
Maximin Coavoux and Beno{\^\i}t Crabb{\'e}.
\newblock 2017.
\newblock Incremental discontinuous phrase structure parsing with the {GAP}
  transition.
\newblock In {\em Proceedings of the 15th Conference of the {E}uropean Chapter
  of the Association for Computational Linguistics: Volume 1, Long Papers},
  pages 1259--1270, Valencia, Spain, April. Association for Computational
  Linguistics.

\bibitem[\protect\citename{Coavoux \bgroup et al.\egroup
  }2019]{coavoux-etal-2019-unlexicalized}
Maximin Coavoux, Beno{\^\i}t Crabb{\'e}, and Shay~B. Cohen.
\newblock 2019.
\newblock Unlexicalized transition-based discontinuous constituency parsing.
\newblock {\em Transactions of the Association for Computational Linguistics},
  7:73--89, March.

\bibitem[\protect\citename{Covington}2001]{covington01fundamental}
Michael~A. Covington.
\newblock 2001.
\newblock A fundamental algorithm for dependency parsing.
\newblock In {\em Proceedings of the 39th Annual ACM Southeast Conference},
  pages 95--102, New York, NY, USA. ACM.

\bibitem[\protect\citename{Crabb{\'e}}2014]{crabbe-2014-lr}
Benoit Crabb{\'e}.
\newblock 2014.
\newblock An {LR}-inspired generalized lexicalized phrase structure parser.
\newblock In {\em Proceedings of {COLING} 2014, the 25th International
  Conference on Computational Linguistics: Technical Papers}, pages 541--552,
  Dublin, Ireland, August. Dublin City University and Association for
  Computational Linguistics.

\bibitem[\protect\citename{Cross and Huang}2016]{cross-huang-2016-span}
James Cross and Liang Huang.
\newblock 2016.
\newblock Span-based constituency parsing with a structure-label system and
  provably optimal dynamic oracles.
\newblock In {\em Proceedings of the 2016 Conference on Empirical Methods in
  Natural Language Processing}, pages 1--11, Austin, Texas, November.
  Association for Computational Linguistics.

\bibitem[\protect\citename{Damonte \bgroup et al.\egroup
  }2017]{damonte-etal-2017-incremental}
Marco Damonte, Shay~B. Cohen, and Giorgio Satta.
\newblock 2017.
\newblock An incremental parser for abstract meaning representation.
\newblock In {\em Proceedings of the 15th Conference of the {E}uropean Chapter
  of the Association for Computational Linguistics: Volume 1, Long Papers},
  pages 536--546, Valencia, Spain, April. Association for Computational
  Linguistics.

\bibitem[\protect\citename{de Lhoneux \bgroup et al.\egroup
  }2017a]{de-lhoneux-etal-2017-raw}
Miryam de~Lhoneux, Yan Shao, Ali Basirat, Eliyahu Kiperwasser, Sara Stymne,
  Yoav Goldberg, and Joakim Nivre.
\newblock 2017a.
\newblock From raw text to universal dependencies - look, no tags!
\newblock In {\em Proceedings of the {C}o{NLL} 2017 Shared Task: Multilingual
  Parsing from Raw Text to Universal Dependencies}, pages 207--217, Vancouver,
  Canada, August. Association for Computational Linguistics.

\bibitem[\protect\citename{de Lhoneux \bgroup et al.\egroup
  }2017b]{de-lhoneux-etal-2017-arc}
Miryam de~Lhoneux, Sara Stymne, and Joakim Nivre.
\newblock 2017b.
\newblock Arc-hybrid non-projective dependency parsing with a static-dynamic
  oracle.
\newblock In {\em Proceedings of the 15th International Conference on Parsing
  Technologies}, pages 99--104, Pisa, Italy, September. Association for
  Computational Linguistics.

\bibitem[\protect\citename{Devlin \bgroup et al.\egroup
  }2019]{devlin-etal-2019-bert}
Jacob Devlin, Ming-Wei Chang, Kenton Lee, and Kristina Toutanova.
\newblock 2019.
\newblock {BERT}: Pre-training of deep bidirectional transformers for language
  understanding.
\newblock In {\em Proceedings of the 2019 Conference of the North {A}merican
  Chapter of the Association for Computational Linguistics: Human Language
  Technologies, Volume 1 (Long and Short Papers)}, pages 4171--4186,
  Minneapolis, Minnesota, June. Association for Computational Linguistics.

\bibitem[\protect\citename{Dyer \bgroup et al.\egroup
  }2016]{dyer-etal-2016-recurrent}
Chris Dyer, Adhiguna Kuncoro, Miguel Ballesteros, and Noah~A. Smith.
\newblock 2016.
\newblock Recurrent neural network grammars.
\newblock In {\em Proceedings of the 2016 Conference of the North {A}merican
  Chapter of the Association for Computational Linguistics: Human Language
  Technologies}, pages 199--209, San Diego, California, June. Association for
  Computational Linguistics.

\bibitem[\protect\citename{Fern{\'a}ndez-Gonz{\'a}lez and
  G{\'o}mez-Rodr{\'\i}guez}2012]{fernandez-gonzalez-gomez-rodriguez-2012-improving}
Daniel Fern{\'a}ndez-Gonz{\'a}lez and Carlos G{\'o}mez-Rodr{\'\i}guez.
\newblock 2012.
\newblock Improving transition-based dependency parsing with buffer
  transitions.
\newblock In {\em Proceedings of the 2012 Joint Conference on Empirical Methods
  in Natural Language Processing and Computational Natural Language Learning},
  pages 308--319, Jeju Island, Korea, July. Association for Computational
  Linguistics.

\bibitem[\protect\citename{Fern{\'a}ndez-Gonz{\'a}lez and
  G{\'o}mez-Rodr{\'\i}guez}2017]{fernandez-gonzalez-gomez-rodriguez-2017-full}
Daniel Fern{\'a}ndez-Gonz{\'a}lez and Carlos G{\'o}mez-Rodr{\'\i}guez.
\newblock 2017.
\newblock A full non-monotonic transition system for unrestricted
  non-projective parsing.
\newblock In {\em Proceedings of the 55th Annual Meeting of the Association for
  Computational Linguistics (Volume 1: Long Papers)}, pages 288--298,
  Vancouver, Canada, July. Association for Computational Linguistics.

\bibitem[\protect\citename{Fern{\'a}ndez-Gonz{\'a}lez and
  G{\'o}mez-Rodr{\'\i}guez}2018]{fernandez-gonzalez-gomez-rodriguez-2018-non}
Daniel Fern{\'a}ndez-Gonz{\'a}lez and Carlos G{\'o}mez-Rodr{\'\i}guez.
\newblock 2018.
\newblock Non-projective dependency parsing with non-local transitions.
\newblock In {\em Proceedings of the 2018 Conference of the North {A}merican
  Chapter of the Association for Computational Linguistics: Human Language
  Technologies, Volume 2 (Short Papers)}, pages 693--700, New Orleans,
  Louisiana, June. Association for Computational Linguistics.

\bibitem[\protect\citename{Fern{\'a}ndez-Gonz{\'a}lez and
  G{\'o}mez-Rodr{\'i}guez}2019a]{FerGomAI2019}
Daniel Fern{\'a}ndez-Gonz{\'a}lez and Carlos G{\'o}mez-Rodr{\'i}guez.
\newblock 2019a.
\newblock Faster shift-reduce constituent parsing with a non-binary, bottom-up
  strategy.
\newblock {\em Artificial Intelligence}, 275:559 -- 574.

\bibitem[\protect\citename{Fern{\'a}ndez-Gonz{\'a}lez and
  G{\'o}mez-Rodr{\'\i}guez}2019b]{fernandez-gonzalez-gomez-rodriguez-2019-left}
Daniel Fern{\'a}ndez-Gonz{\'a}lez and Carlos G{\'o}mez-Rodr{\'\i}guez.
\newblock 2019b.
\newblock Left-to-right dependency parsing with pointer networks.
\newblock In {\em Proceedings of the 2019 Conference of the North {A}merican
  Chapter of the Association for Computational Linguistics: Human Language
  Technologies, Volume 1 (Long and Short Papers)}, pages 710--716, Minneapolis,
  Minnesota, June. Association for Computational Linguistics.

\bibitem[\protect\citename{Fraser}1989]{fraser1989parsing}
Norman Fraser.
\newblock 1989.
\newblock Parsing and dependency grammar.
\newblock {\em Ucl Working Papers in Linguistics 1: University College London},
  pages 296--319.

\bibitem[\protect\citename{Ginter \bgroup et al.\egroup
  }2017]{ginter-et-al-2017-conll}
Filip Ginter, Jan Haji{\v c}, Juhani Luotolahti, Milan Straka, and Daniel
  Zeman.
\newblock 2017.
\newblock {CoNLL} 2017 shared task - automatically annotated raw texts and word
  embeddings.
\newblock {LINDAT}/{CLARIN} digital library at the Institute of Formal and
  Applied Linguistics ({{\'U}FAL}), Faculty of Mathematics and Physics, Charles
  University.

\bibitem[\protect\citename{Goldberg and
  Elhadad}2010]{goldberg-elhadad-2010-efficient}
Yoav Goldberg and Michael Elhadad.
\newblock 2010.
\newblock An efficient algorithm for easy-first non-directional dependency
  parsing.
\newblock In {\em Human Language Technologies: The 2010 Annual Conference of
  the North {A}merican Chapter of the Association for Computational
  Linguistics}, pages 742--750, Los Angeles, California, June. Association for
  Computational Linguistics.

\bibitem[\protect\citename{Goldberg and
  Nivre}2012]{goldberg-nivre-2012-dynamic}
Yoav Goldberg and Joakim Nivre.
\newblock 2012.
\newblock A dynamic oracle for arc-eager dependency parsing.
\newblock In {\em Proceedings of {COLING} 2012}, pages 959--976, Mumbai, India,
  December. The COLING 2012 Organizing Committee.

\bibitem[\protect\citename{G{\'o}mez-Rodr{\'\i}guez and
  Nivre}2010]{gomez-rodriguez-nivre-2010-transition}
Carlos G{\'o}mez-Rodr{\'\i}guez and Joakim Nivre.
\newblock 2010.
\newblock A transition-based parser for 2-planar dependency structures.
\newblock In {\em Proceedings of the 48th Annual Meeting of the Association for
  Computational Linguistics}, pages 1492--1501, Uppsala, Sweden, July.
  Association for Computational Linguistics.

\bibitem[\protect\citename{G{\'o}mez-Rodr{\'\i}guez and
  Vilares}2018]{gomez-rodriguez-vilares-2018-constituent}
Carlos G{\'o}mez-Rodr{\'\i}guez and David Vilares.
\newblock 2018.
\newblock Constituent parsing as sequence labeling.
\newblock In {\em Proceedings of the 2018 Conference on Empirical Methods in
  Natural Language Processing}, pages 1314--1324, Brussels, Belgium,
  October-November. Association for Computational Linguistics.

\bibitem[\protect\citename{Hershcovich \bgroup et al.\egroup
  }2017]{hershcovich-etal-2017-transition}
Daniel Hershcovich, Omri Abend, and Ari Rappoport.
\newblock 2017.
\newblock A transition-based directed acyclic graph parser for {UCCA}.
\newblock In {\em Proceedings of the 55th Annual Meeting of the Association for
  Computational Linguistics (Volume 1: Long Papers)}, pages 1127--1138,
  Vancouver, Canada, July. Association for Computational Linguistics.

\bibitem[\protect\citename{Hershcovich \bgroup et al.\egroup
  }2018]{hershcovich-etal-2018-multitask}
Daniel Hershcovich, Omri Abend, and Ari Rappoport.
\newblock 2018.
\newblock Multitask parsing across semantic representations.
\newblock In {\em Proceedings of the 56th Annual Meeting of the Association for
  Computational Linguistics (Volume 1: Long Papers)}, pages 373--385,
  Melbourne, Australia, July. Association for Computational Linguistics.

\bibitem[\protect\citename{Hochreiter and Schmidhuber}1997]{hochreiter1997long}
Sepp Hochreiter and J{\"u}rgen Schmidhuber.
\newblock 1997.
\newblock Long short-term memory.
\newblock {\em Neural computation}, 9(8):1735--1780.

\bibitem[\protect\citename{Honnibal and
  Johnson}2015]{honnibal-johnson-2015-improved}
Matthew Honnibal and Mark Johnson.
\newblock 2015.
\newblock An improved non-monotonic transition system for dependency parsing.
\newblock In {\em Proceedings of the 2015 Conference on Empirical Methods in
  Natural Language Processing}, pages 1373--1378, Lisbon, Portugal, September.
  Association for Computational Linguistics.

\bibitem[\protect\citename{Honnibal \bgroup et al.\egroup
  }2013]{honnibal-etal-2013-non}
Matthew Honnibal, Yoav Goldberg, and Mark Johnson.
\newblock 2013.
\newblock A non-monotonic arc-eager transition system for dependency parsing.
\newblock In {\em Proceedings of the Seventeenth Conference on Computational
  Natural Language Learning}, pages 163--172, Sofia, Bulgaria, August.
  Association for Computational Linguistics.

\bibitem[\protect\citename{Kiperwasser and
  Goldberg}2016]{kiperwasser-goldberg-2016-simple}
Eliyahu Kiperwasser and Yoav Goldberg.
\newblock 2016.
\newblock Simple and accurate dependency parsing using bidirectional {LSTM}
  feature representations.
\newblock {\em Transactions of the Association for Computational Linguistics},
  4:313--327.

\bibitem[\protect\citename{Kitaev and Klein}2019]{tetra}
Nikita Kitaev and Dan Klein.
\newblock 2019.
\newblock Tetra-tagging: Word-synchronous parsing with linear-time inference.
\newblock {\em CoRR}, abs/1904.09745.

\bibitem[\protect\citename{Konstas \bgroup et al.\egroup
  }2017]{konstas-etal-2017-neural}
Ioannis Konstas, Srinivasan Iyer, Mark Yatskar, Yejin Choi, and Luke
  Zettlemoyer.
\newblock 2017.
\newblock Neural {AMR}: Sequence-to-sequence models for parsing and generation.
\newblock In {\em Proceedings of the 55th Annual Meeting of the Association for
  Computational Linguistics (Volume 1: Long Papers)}, pages 146--157,
  Vancouver, Canada, July. Association for Computational Linguistics.

\bibitem[\protect\citename{Kuhlmann \bgroup et al.\egroup
  }2011]{kuhlmann-etal-2011-dynamic}
Marco Kuhlmann, Carlos G{\'o}mez-Rodr{\'\i}guez, and Giorgio Satta.
\newblock 2011.
\newblock Dynamic programming algorithms for transition-based dependency
  parsers.
\newblock In {\em Proceedings of the 49th Annual Meeting of the Association for
  Computational Linguistics: Human Language Technologies}, pages 673--682,
  Portland, Oregon, USA, June. Association for Computational Linguistics.

\bibitem[\protect\citename{Li \bgroup et al.\egroup
  }2018]{li-etal-2018-seq2seq}
Zuchao Li, Jiaxun Cai, Shexia He, and Hai Zhao.
\newblock 2018.
\newblock Seq2seq dependency parsing.
\newblock In {\em Proceedings of the 27th International Conference on
  Computational Linguistics}, pages 3203--3214, Santa Fe, New Mexico, USA,
  August. Association for Computational Linguistics.

\bibitem[\protect\citename{Liu and Zhang}2017a]{liu-zhang-2017-encoder}
Jiangming Liu and Yue Zhang.
\newblock 2017a.
\newblock Encoder-decoder shift-reduce syntactic parsing.
\newblock In {\em Proceedings of the 15th International Conference on Parsing
  Technologies}, pages 105--114, Pisa, Italy, September. Association for
  Computational Linguistics.

\bibitem[\protect\citename{Liu and Zhang}2017b]{liu-zhang-2017-order}
Jiangming Liu and Yue Zhang.
\newblock 2017b.
\newblock In-order transition-based constituent parsing.
\newblock {\em Transactions of the Association for Computational Linguistics},
  5:413--424.

\bibitem[\protect\citename{Ma \bgroup et al.\egroup }2018]{ma-etal-2018-stack}
Xuezhe Ma, Zecong Hu, Jingzhou Liu, Nanyun Peng, Graham Neubig, and Eduard
  Hovy.
\newblock 2018.
\newblock Stack-pointer networks for dependency parsing.
\newblock In {\em Proceedings of the 56th Annual Meeting of the Association for
  Computational Linguistics (Volume 1: Long Papers)}, pages 1403--1414,
  Melbourne, Australia, July. Association for Computational Linguistics.

\bibitem[\protect\citename{Maier and
  Lichte}2016]{maier-lichte-2016-discontinuous}
Wolfgang Maier and Timm Lichte.
\newblock 2016.
\newblock Discontinuous parsing with continuous trees.
\newblock In {\em Proceedings of the Workshop on Discontinuous Structures in
  Natural Language Processing}, pages 47--57, San Diego, California, June.
  Association for Computational Linguistics.

\bibitem[\protect\citename{Maier}2015]{maier-2015-discontinuous}
Wolfgang Maier.
\newblock 2015.
\newblock Discontinuous incremental shift-reduce parsing.
\newblock In {\em Proceedings of the 53rd Annual Meeting of the Association for
  Computational Linguistics and the 7th International Joint Conference on
  Natural Language Processing (Volume 1: Long Papers)}, pages 1202--1212,
  Beijing, China, July. Association for Computational Linguistics.

\bibitem[\protect\citename{Mi and Huang}2015]{mi-huang-2015-shift}
Haitao Mi and Liang Huang.
\newblock 2015.
\newblock Shift-reduce constituency parsing with dynamic programming and {POS}
  tag lattice.
\newblock In {\em Proceedings of the 2015 Conference of the North {A}merican
  Chapter of the Association for Computational Linguistics: Human Language
  Technologies}, pages 1030--1035, Denver, Colorado, May{--}June. Association
  for Computational Linguistics.

\bibitem[\protect\citename{Nivre and
  Fern{\'a}ndez-Gonz{\'a}lez}2014]{nivre-fernandez-gonzalez-2014-squibs}
Joakim Nivre and Daniel Fern{\'a}ndez-Gonz{\'a}lez.
\newblock 2014.
\newblock {S}quibs: Arc-eager parsing with the tree constraint.
\newblock {\em Computational Linguistics}, 40(2):259--267, June.

\bibitem[\protect\citename{Nivre and others}2019]{ud2.4}
Joakim Nivre et~al.
\newblock 2019.
\newblock Universal dependencies 2.4.
\newblock {LINDAT}/{CLARIAH}-{CZ} digital library at the Institute of Formal
  and Applied Linguistics ({{\'U}FAL}), Faculty of Mathematics and Physics,
  Charles University.

\bibitem[\protect\citename{Nivre \bgroup et al.\egroup
  }2004]{nivre-etal-2004-memory}
Joakim Nivre, Johan Hall, and Jens Nilsson.
\newblock 2004.
\newblock Memory-based dependency parsing.
\newblock In {\em Proceedings of the Eighth Conference on Computational Natural
  Language Learning ({C}o{NLL}-2004) at {HLT}-{NAACL} 2004}, pages 49--56,
  Boston, Massachusetts, USA, May 6 - May 7. Association for Computational
  Linguistics.

\bibitem[\protect\citename{Nivre}2003]{nivre-2003-efficient}
Joakim Nivre.
\newblock 2003.
\newblock An efficient algorithm for projective dependency parsing.
\newblock In {\em Proceedings of the Eighth International Conference on Parsing
  Technologies}, pages 149--160, Nancy, France, April.

\bibitem[\protect\citename{Nivre}2004]{nivre-2004-incrementality}
Joakim Nivre.
\newblock 2004.
\newblock Incrementality in deterministic dependency parsing.
\newblock In {\em Proceedings of the Workshop on Incremental Parsing: Bringing
  Engineering and Cognition Together}, pages 50--57, Barcelona, Spain, July.
  Association for Computational Linguistics.

\bibitem[\protect\citename{Nivre}2008]{nivre-2008-algorithms}
Joakim Nivre.
\newblock 2008.
\newblock Algorithms for deterministic incremental dependency parsing.
\newblock {\em Computational Linguistics}, 34(4):513--553.

\bibitem[\protect\citename{Nivre}2009]{nivre-2009-non}
Joakim Nivre.
\newblock 2009.
\newblock Non-projective dependency parsing in expected linear time.
\newblock In {\em Proceedings of the Joint Conference of the 47th Annual
  Meeting of the {ACL} and the 4th International Joint Conference on Natural
  Language Processing of the {AFNLP}}, pages 351--359, Suntec, Singapore,
  August. Association for Computational Linguistics.

\bibitem[\protect\citename{Pitler and
  McDonald}2015]{pitler-mcdonald-2015-linear}
Emily Pitler and Ryan McDonald.
\newblock 2015.
\newblock A linear-time transition system for crossing interval trees.
\newblock In {\em Proceedings of the 2015 Conference of the North {A}merican
  Chapter of the Association for Computational Linguistics: Human Language
  Technologies}, pages 662--671, Denver, Colorado, May{--}June. Association for
  Computational Linguistics.

\bibitem[\protect\citename{Qi and Manning}2017]{qi-manning-2017-arc}
Peng Qi and Christopher~D. Manning.
\newblock 2017.
\newblock Arc-swift: A novel transition system for dependency parsing.
\newblock In {\em Proceedings of the 55th Annual Meeting of the Association for
  Computational Linguistics (Volume 2: Short Papers)}, pages 110--117,
  Vancouver, Canada, July. Association for Computational Linguistics.

\bibitem[\protect\citename{Ribeyre \bgroup et al.\egroup
  }2014]{ribeyre-etal-2014-alpage}
Corentin Ribeyre, Eric Villemonte de~la Clergerie, and Djam{\'e} Seddah.
\newblock 2014.
\newblock {A}lpage: Transition-based semantic graph parsing with syntactic
  features.
\newblock In {\em Proceedings of the 8th International Workshop on Semantic
  Evaluation ({S}em{E}val 2014)}, pages 97--103, Dublin, Ireland, August.
  Association for Computational Linguistics.

\bibitem[\protect\citename{Sagae and Lavie}2005]{sagae-lavie-2005-classifier}
Kenji Sagae and Alon Lavie.
\newblock 2005.
\newblock A classifier-based parser with linear run-time complexity.
\newblock In {\em Proceedings of the Ninth International Workshop on Parsing
  Technology}, pages 125--132, Vancouver, British Columbia, October.
  Association for Computational Linguistics.

\bibitem[\protect\citename{Sagae and Tsujii}2008]{sagae-tsujii-2008-shift}
Kenji Sagae and Jun{'}ichi Tsujii.
\newblock 2008.
\newblock Shift-reduce dependency {DAG} parsing.
\newblock In {\em Proceedings of the 22nd International Conference on
  Computational Linguistics (Coling 2008)}, pages 753--760, Manchester, UK,
  August. Coling 2008 Organizing Committee.

\bibitem[\protect\citename{Schuster and
  Paliwal}1997]{schuster1997bidirectional}
Mike Schuster and Kuldip~K Paliwal.
\newblock 1997.
\newblock Bidirectional recurrent neural networks.
\newblock {\em IEEE transactions on Signal Processing}, 45(11):2673--2681.

\bibitem[\protect\citename{Shi \bgroup et al.\egroup }2017]{shi-etal-2017-fast}
Tianze Shi, Liang Huang, and Lillian Lee.
\newblock 2017.
\newblock Fast(er) exact decoding and global training for transition-based
  dependency parsing via a minimal feature set.
\newblock In {\em Proceedings of the 2017 Conference on Empirical Methods in
  Natural Language Processing}, pages 12--23, Copenhagen, Denmark, September.
  Association for Computational Linguistics.

\bibitem[\protect\citename{Smith \bgroup et al.\egroup
  }2018]{smith-etal-2018-investigation}
Aaron Smith, Miryam de~Lhoneux, Sara Stymne, and Joakim Nivre.
\newblock 2018.
\newblock An investigation of the interactions between pre-trained word
  embeddings, character models and {POS} tags in dependency parsing.
\newblock In {\em Proceedings of the 2018 Conference on Empirical Methods in
  Natural Language Processing}, pages 2711--2720, Brussels, Belgium,
  October-November. Association for Computational Linguistics.

\bibitem[\protect\citename{Spoustov{\'a} and
  Spousta}2010]{spoustova2010dependency}
Drahom{\'\i}ra Spoustov{\'a} and Miroslav Spousta.
\newblock 2010.
\newblock Dependency parsing as a sequence labeling task.
\newblock {\em The Prague Bulletin of Mathematical Linguistics},
  94(2010):7--14.

\bibitem[\protect\citename{Stanojevi{\'c} and
  G.~Alhama}2017]{stanojevic-g-alhama-2017-neural}
Milo{\v{s}} Stanojevi{\'c} and Raquel G.~Alhama.
\newblock 2017.
\newblock Neural discontinuous constituency parsing.
\newblock In {\em Proceedings of the 2017 Conference on Empirical Methods in
  Natural Language Processing}, pages 1666--1676, Copenhagen, Denmark,
  September. Association for Computational Linguistics.

\bibitem[\protect\citename{Straka}2018]{straka-2018-udpipe}
Milan Straka.
\newblock 2018.
\newblock {UDP}ipe 2.0 prototype at {C}o{NLL} 2018 {UD} shared task.
\newblock In {\em Proceedings of the {C}o{NLL} 2018 Shared Task: Multilingual
  Parsing from Raw Text to Universal Dependencies}, pages 197--207, Brussels,
  Belgium, October. Association for Computational Linguistics.

\bibitem[\protect\citename{Strzyz \bgroup et al.\egroup
  }2019]{strzyz-etal-2019-viable}
Michalina Strzyz, David Vilares, and Carlos G{\'o}mez-Rodr{\'\i}guez.
\newblock 2019.
\newblock Viable dependency parsing as sequence labeling.
\newblock In {\em Proceedings of the 2019 Conference of the North {A}merican
  Chapter of the Association for Computational Linguistics: Human Language
  Technologies, Volume 1 (Long and Short Papers)}, pages 717--723, Minneapolis,
  Minnesota, June. Association for Computational Linguistics.

\bibitem[\protect\citename{Swayamdipta \bgroup et al.\egroup
  }2016]{swayamdipta-etal-2016-greedy}
Swabha Swayamdipta, Miguel Ballesteros, Chris Dyer, and Noah~A. Smith.
\newblock 2016.
\newblock Greedy, joint syntactic-semantic parsing with stack {LSTM}s.
\newblock In {\em Proceedings of The 20th {SIGNLL} Conference on Computational
  Natural Language Learning}, pages 187--197, Berlin, Germany, August.
  Association for Computational Linguistics.

\bibitem[\protect\citename{Titov \bgroup et al.\egroup }2009]{titov2009online}
Ivan Titov, James Henderson, Paola Merlo, and Gabriele Musillo.
\newblock 2009.
\newblock Online graph planarisation for synchronous parsing of semantic and
  syntactic dependencies.
\newblock {\em IJCAI}, pages 1562--1567.

\bibitem[\protect\citename{Tokg{\"o}z and
  Eryi{\v{g}}it}2015]{tokgoz-eryigit-2015-transition}
Alper Tokg{\"o}z and G{\"u}l{\c{s}}en Eryi{\v{g}}it.
\newblock 2015.
\newblock Transition-based dependency {DAG} parsing using dynamic oracles.
\newblock In {\em Proceedings of the {ACL}-{IJCNLP} 2015 Student Research
  Workshop}, pages 22--27, Beijing, China, July. Association for Computational
  Linguistics.

\bibitem[\protect\citename{Vacareanu \bgroup et al.\egroup
  }2020]{vacareanu-etal-2020-parsing}
Robert Vacareanu, George~Caique Gouveia~Barbosa, Marco~A.
  Valenzuela-Esc{\'a}rcega, and Mihai Surdeanu.
\newblock 2020.
\newblock Parsing as tagging.
\newblock In {\em Proceedings of The 12th Language Resources and Evaluation
  Conference}, pages 5225--5231, Marseille, France, May. European Language
  Resources Association.

\bibitem[\protect\citename{Versley}2014]{versley-2014-experiments}
Yannick Versley.
\newblock 2014.
\newblock Experiments with easy-first nonprojective constituent parsing.
\newblock In {\em Proceedings of the First Joint Workshop on Statistical
  Parsing of Morphologically Rich Languages and Syntactic Analysis of
  Non-Canonical Languages}, pages 39--53, Dublin, Ireland, August. Dublin City
  University.

\bibitem[\protect\citename{Vilares and
  G{\'o}mez-Rodr{\'\i}guez}2018]{vilares-gomez-rodriguez-2018-transition}
David Vilares and Carlos G{\'o}mez-Rodr{\'\i}guez.
\newblock 2018.
\newblock Transition-based parsing with lighter feed-forward networks.
\newblock In {\em Proceedings of the Second Workshop on Universal Dependencies
  ({UDW} 2018)}, pages 162--172, Brussels, Belgium, November. Association for
  Computational Linguistics.

\bibitem[\protect\citename{Vinyals \bgroup et al.\egroup }2015]{Vinyals2015}
Oriol Vinyals, Lukasz Kaiser, Terry Koo, Slav Petrov, Ilya Sutskever, and
  Geoffrey Hinton.
\newblock 2015.
\newblock Grammar as a foreign language.
\newblock In {\em Proceedings of the 28th International Conference on Neural
  Information Processing Systems - Volume 2}, NIPS'15, pages 2773--2781,
  Cambridge, MA, USA. MIT Press.

\bibitem[\protect\citename{Virtanen \bgroup et al.\egroup
  }2019]{virtanen2019multilingual}
Antti Virtanen, Jenna Kanerva, Rami Ilo, Jouni Luoma, Juhani Luotolahti, Tapio
  Salakoski, Filip Ginter, and Sampo Pyysalo.
\newblock 2019.
\newblock Multilingual is not enough: Bert for finnish.
\newblock {\em arXiv preprint arXiv:1912.07076}.

\bibitem[\protect\citename{Wang \bgroup et al.\egroup
  }2015]{wang-etal-2015-transition}
Chuan Wang, Nianwen Xue, and Sameer Pradhan.
\newblock 2015.
\newblock A transition-based algorithm for {AMR} parsing.
\newblock In {\em Proceedings of the 2015 Conference of the North {A}merican
  Chapter of the Association for Computational Linguistics: Human Language
  Technologies}, pages 366--375, Denver, Colorado, May{--}June. Association for
  Computational Linguistics.

\bibitem[\protect\citename{Wang \bgroup et al.\egroup }2018]{Wang2018}
Yuxuan Wang, Wanxiang Che, Jiang Guo, and Ting Liu.
\newblock 2018.
\newblock A neural transition-based approach for semantic dependency graph
  parsing.

\bibitem[\protect\citename{Wu \bgroup et al.\egroup }2016]{wu2016google}
Yonghui Wu, Mike Schuster, Zhifeng Chen, Quoc~V Le, Mohammad Norouzi, Wolfgang
  Macherey, Maxim Krikun, Yuan Cao, Qin Gao, Klaus Macherey, et~al.
\newblock 2016.
\newblock Google's neural machine translation system: Bridging the gap between
  human and machine translation.
\newblock {\em arXiv preprint arXiv:1609.08144}.

\bibitem[\protect\citename{Yamada and
  Matsumoto}2003]{yamada-matsumoto-2003-statistical}
Hiroyasu Yamada and Yuji Matsumoto.
\newblock 2003.
\newblock Statistical dependency analysis with support vector machines.
\newblock In {\em Proceedings of the Eighth International Conference on Parsing
  Technologies}, pages 195--206, Nancy, France, April.

\bibitem[\protect\citename{Yang and Zhang}2018]{yang-zhang-2018-ncrf}
Jie Yang and Yue Zhang.
\newblock 2018.
\newblock {NCRF}++: An open-source neural sequence labeling toolkit.
\newblock In {\em Proceedings of {ACL} 2018, System Demonstrations}, pages
  74--79, Melbourne, Australia, July. Association for Computational
  Linguistics.

\bibitem[\protect\citename{Yli-Jyr{\"a} and
  G{\'o}mez-Rodr{\'\i}guez}2017]{yli-jyra-gomez-rodriguez-2017-generic}
Anssi Yli-Jyr{\"a} and Carlos G{\'o}mez-Rodr{\'\i}guez.
\newblock 2017.
\newblock Generic axiomatization of families of noncrossing graphs in
  dependency parsing.
\newblock In {\em Proceedings of the 55th Annual Meeting of the Association for
  Computational Linguistics (Volume 1: Long Papers)}, pages 1745--1755,
  Vancouver, Canada, July. Association for Computational Linguistics.

\bibitem[\protect\citename{Zhang and Clark}2009]{zhang-clark-2009-transition}
Yue Zhang and Stephen Clark.
\newblock 2009.
\newblock Transition-based parsing of the {C}hinese treebank using a global
  discriminative model.
\newblock In {\em Proceedings of the 11th International Conference on Parsing
  Technologies ({IWPT}{'}09)}, pages 162--171, Paris, France, October.
  Association for Computational Linguistics.

\bibitem[\protect\citename{Zhang and Nivre}2011]{zhang-nivre-2011-transition}
Yue Zhang and Joakim Nivre.
\newblock 2011.
\newblock Transition-based dependency parsing with rich non-local features.
\newblock In {\em Proceedings of the 49th Annual Meeting of the Association for
  Computational Linguistics: Human Language Technologies}, pages 188--193,
  Portland, Oregon, USA, June. Association for Computational Linguistics.

\bibitem[\protect\citename{Zhang \bgroup et al.\egroup
  }2016]{zhang-etal-2016-transition}
Meishan Zhang, Yue Zhang, and Guohong Fu.
\newblock 2016.
\newblock Transition-based neural word segmentation.
\newblock In {\em Proceedings of the 54th Annual Meeting of the Association for
  Computational Linguistics (Volume 1: Long Papers)}, pages 421--431, Berlin,
  Germany, August. Association for Computational Linguistics.

\bibitem[\protect\citename{Zhang \bgroup et al.\egroup
  }2017]{zhang-etal-2017-stack}
Zhirui Zhang, Shujie Liu, Mu~Li, Ming Zhou, and Enhong Chen.
\newblock 2017.
\newblock Stack-based multi-layer attention for transition-based dependency
  parsing.
\newblock In {\em Proceedings of the 2017 Conference on Empirical Methods in
  Natural Language Processing}, pages 1677--1682, Copenhagen, Denmark,
  September. Association for Computational Linguistics.

\bibitem[\protect\citename{Zhu \bgroup et al.\egroup }2013]{zhu-etal-2013-fast}
Muhua Zhu, Yue Zhang, Wenliang Chen, Min Zhang, and Jingbo Zhu.
\newblock 2013.
\newblock Fast and accurate shift-reduce constituent parsing.
\newblock In {\em Proceedings of the 51st Annual Meeting of the Association for
  Computational Linguistics (Volume 1: Long Papers)}, pages 434--443, Sofia,
  Bulgaria, August. Association for Computational Linguistics.

\end{thebibliography}

\clearpage
\appendix

\section{Label sizes}\label{appendix-label-sizes}

\begin{table}[hbtp!]
\centering
\small{
\scalebox{0.78}{
\begin{tabular}{llccllcc}
\toprule
\multirow{3}{*}{Treebank} & \multirow{3}{*}{Encoding} & Task 1 \#labels & Task 2 \#labels &  \multirow{3}{*}{Treebank} & \multirow{3}{*}{Encoding} & Task 1 \#labels & Task 2 \#labels \\
&&(subsequence&(dependency&&&(subsequence&(dependency\\
&&of transitions)&relations)&&&of transitions)&relations)\\ \midrule
\multirow{6}{*}{Ancient Greek} & rel-pos &166 &27 & \multirow{6}{*}{Russian} & rel-pos & 169 & 44 \\
 & bracketing & 210	&27 & & bracketing & 93 & 44 \\ 
 & arc-standard & 46& 27 &  & arc-standard & 65 & 44 \\
 & arc-eager &501&27 &  & arc-eager & 385 & 44 \\
 & arc-hybrid &41&	27 & & arc-hybrid & 45 & 44 \\ 
 & covington & 3651& 27 &  & covington & 1573 & 44 \\ \midrule
\multirow{6}{*}{Chinese} & rel-pos & 201 & 45 & \multirow{6}{*}{Tamil} & rel-pos & 68 & 31 \\
 & bracketing & 104 & 45 & & bracketing & 49 & 31 \\ 
 & arc-standard & 56 & 45 &  & arc-standard & 22 & 31 \\
 & arc-eager & 817 & 45 &  & arc-eager & 79 & 31 \\
 & arc-hybrid & 54 & 45 &  & arc-hybrid & 25 & 31 \\
 & covington & 4846 & 45 &  & covington & 543 & 31 \\ \midrule
\multirow{6}{*}{English} & rel-pos & 202 & 51 & \multirow{6}{*}{Uyghur} & rel-pos &102 &44 \\ 
 & bracketing & 109 & 51 &  & bracketing & 58	&44 \\
 & arc-standard & 90 & 51 &  & arc-standard & 20	&44 \\
 & arc-eager & 509 & 51 & &  arc-eager & 185&	44 \\
 & arc-hybrid & 52 & 51 &  & arc-hybrid & 26&	44\\
 & covington & 2804 & 51  & & covington & 1459&	44\\ \midrule
\multirow{6}{*}{Finnish} & rel-pos & 220 & 47 &  \multirow{6}{*}{Wolof} & rel-pos &102&	40\\
 & bracketing & 107 & 47 &  & bracketing & 69&	40\\
 & arc-standard & 77 & 47 &  & arc-standard & 69&	40 \\
 & arc-eager & 356 & 47 &  & arc-eager & 373&	40\\
 & arc-hybrid & 46 & 47 &  & arc-hybrid & 46&	40 \\
 & covington & 2185 & 47 & &  covington & 920&	40 \\ \midrule
\multirow{6}{*}{Hebrew} & rel-pos & 162 & 40 \\
 & bracketing & 107 & 40 \\
 & arc-standard & 76 & 40 \\
 & arc-hybrid & 58 & 40 \\
 & covington & 2070 & 40 \\ \bottomrule
\end{tabular}
}}
\caption{Number of labels for the transition-based parsing algorithms (arc-standard, arc-eager, arc-hybrid and Covington). We include two other existing encodings coming from different families (unrelated to transition-based parsing): the rel-PoS and bracketing encodings \cite{strzyz-etal-2019-viable}.}
\label{tab:label-size}
\end{table}

In Table \ref{tab:label-size} we show the number of distinct labels that several transition-based algorithms generate when mapped to a sequence labeling setup, according to our theory, on the UD treebanks used in our experiments. For comparison purposes, we also include the number of labels of other existing encodings for sequence labeling dependency parsing presented in \newcite{strzyz-etal-2019-viable}: the rel-PoS encoding \cite{spoustova2010dependency} and the bracketing encoding \cite{yli-jyra-gomez-rodriguez-2017-generic}.

\section{Model parameters}\label{appendix-hyperparameters}

\paragraph{BiLSTMs (NCRF++)}

All models were trained with learning rate $0.02$ and learning rate decay of $0.05$, for $100$ epochs with batch size $8$ for training and $128$ for testing. The dimension of the pretrained word vectors was $100$, and $30$ for character embeddings. If PoS tag embeddings were used, their dimension was $25$. The word/character hidden vector dimension was set to $800$ and $50$, respectively. We used $2$ BiLSTM layers with momentum $0.9$. Dependency labels were learned in a multitask learning setup.

\paragraph{BERT}

All models were fine-tuned with the learning rate $10^{-5}$ for $45$ epochs with batch size $8$. The maximum sequence length was set to $400$, except for Russian, with $510$. 
\section{Postprocessing}\label{appendix-postprocessing}

In case a model outputs a label with an illegal action (due to violating preconditions in the transition system, e.g. a left arc in the arc-eager algorithm when the topmost stack node already has a head), we discard it and move to the next predicted action in the sequence. 

The transition systems we implement guarantee acyclicity and single-headedness, but do not guarantee the output to have a single root as required in UD, so we perform simple postprocessing for this purpose. In the implementation of the transition systems, we distinguish between nodes that have not been assigned a head and nodes that have been assigned the dummy root, 0, explicitly as head. With this, it can happen that a tree is generated with no roots (due to headless nodes). In this case all words with the predicted dependency relation ``root'' are assigned 0 as head, i.e. taken as candidates for being the true syntactic root. If still no root has been found, we choose the first token. If there are multiple roots, we pick the first token among them as root and link the others as dependents. Finally, any remaining headless nodes are assigned as dependents of the syntactic root (i.e. the node whose head is the dummy root).

\section{Encodings' upper bound on gold datasets}\label{upperbounds}

Table \ref{tab:gold} reports the performance (UAS and LAS) for the nine treebanks when our BiLSTM models are trained with gold segmentation, tokenization and PoS tags, which serves as an empirical upper bound. In addition, the percentage increase when compared with the predicted setup (using UDpipe) is shown in parentheses. For orientation purposes, we provide the accuracy of UDpipe 1 on the UDv2.4 test sets in Table \ref{tab:udpipe}; for segmentation, tokenization and PoS tagging.

\newcolumntype{g}{>{\columncolor{Gray}}c}
\begin{table*}[hptb!]
\centering
\small{
\tabcolsep=0.065cm
\begin{tabular}{l|ggccgg}
\toprule
\multirow{3}{*}{Encoding}&\multicolumn{2}{g}{Ancient Greek}&\multicolumn{2}{c}{Chinese}&\multicolumn{2}{g}{English}\\
&\tiny{UAS}&\tiny{LAS}&\tiny{UAS}&\tiny{LAS}&\tiny{UAS}&\tiny{LAS}\\
\hline
rel-PoS &74.8 \textit{(+7.6\%)}&	68.3	\textit{(+13.9\%)}&	84.9	\textit{(+31.8\%)}&	82.0	\textit{(+36.7\%)}&	89.3	\textit{(+8.9\%)}&	86.9	\textit{(+10.7\%)}	 \\
Brackets  & 69.1	\textit{(+6.4\%)}&	63.1	\textit{(+12.3\%)}&	84.2	\textit{(+31.1\%)}&	81.2	\textit{(+35.8\%)}&	88.8 \textit{(+8.7\%)}&	86.5	\textit{(+10.4\%)}	 \\
Standard&  65.1	\textit{(+6.7\%)}&	59.6 \textit{(+13.4\%)}&	82.8	\textit{(+31.6\%)}&	79.9	\textit{(+36.6\%)}&	85.7 \textit{(+8.2\%)}&	83.4	\textit{(+10.0\%)}	 \\
Eager  & 64.3	\textit{(+6.4\%)}&	58.6 \textit{(+12.6\%)}&	83.7	\textit{(+32.4\%)}&	81.1	\textit{(+37.4\%)}&	86.6 \textit{(+8.0\%)}&	84.2	\textit{(+9.8\%)}	 \\
Hybrid & 66.2	\textit{(+6.6\%)}&	60.3	\textit{(+12.7\%)}&	83.0 \textit{(+31.4\%)}&	80.1	\textit{(+36.2\%)}&	86.5	\textit{(+8.8\%)}&	84.2	\textit{(+10.7\%)}	 \\
Covington & 67.9	\textit{(+6.6\%)}&	61.6	\textit{(+13.2\%)}&	76.3	\textit{(+30.8\%)}&	73.4	\textit{(+35.4\%)}&	85.1	\textit{(+8.5\%)}&	82.8	\textit{(+10.4\%)}	 \\\midrule
\textit{avg} &  \textit{(+6.7\%)}&		\textit{(+13.0\%)}&		\textbf{\textit{(+31.5\%)}}&		\textbf{\textit{(+36.4\%)}}&		\textit{(+8.5\%)}&		\textit{(+10.3\%)}	 \\ \midrule
\multirow{3}{*}{Encoding}&\multicolumn{2}{g}{Finnish}&\multicolumn{2}{c}{Hebrew}&\multicolumn{2}{g}{Russian}\\

&\tiny{UAS}&\tiny{LAS}&\tiny{UAS}&\tiny{LAS}&\tiny{UAS}&\tiny{LAS}\\
\hline
rel-PoS &85.1	\textit{(+6.2\%)}&	81.3 \textit{(+8.5\%)}&	88.2	\textit{(+40.1\%)}&	85.4	\textit{(+43.7\%)}&	86.3	\textit{(+4.3\%)}&	81.5	\textit{(+5.3\%)}	 \\
Brackets  & 86.1	\textit{(+6.2\%)}&	82.5	\textit{(+8.5\%)}&	87.5	\textit{(+40.9\%)}&	84.7	\textit{(+44.3\%)}&	86.0	\textit{(+3.7\%)}&	81.6	\textit{(+4.8\%)}	 \\
Standard & 83.1	\textit{(+5.9\%)}&	79.6	\textit{(+8.5\%)}&	84.2	\textit{(+39.7\%)}&	81.4	\textit{(+43.9\%)}&	82.2	\textit{(+3.9\%)}&	77.9	\textit{(+4.9\%)}	 \\
Eager  & 83.9	\textit{(+6.2\%)}&	80.0	\textit{(+8.6\%)}&	85.6	\textit{(+39.8\%)}&	82.6	\textit{(+43.6\%)}&	83.7	\textit{(+4.1\%)}&	79.2	\textit{(+5.2\%)}	 \\
Hybrid  & 83.5	\textit{(+5.8\%)}&	80.1	\textit{(+8.2\%)}&	85.1	\textit{(+40.9\%)} &	82.2	\textit{(+44.8\%)}&	82.4	\textit{(+4.0\%)}&	78.1	\textit{(+5.0\%)}	 \\
Covington  & 83.5	\textit{(+5.8\%)}&	79.7	\textit{(+8.1\%)}&	84.0	\textit{(+39.5\%)}&	81.1	\textit{(+43.3\%)}&	82.2	\textit{(+3.8\%)}&	77.8	\textit{(+5.1\%)}	 \\ \midrule
\textit{avg}  & \textit{(+6.0\%)}&		\textit{(+8.4\%)}&		\textbf{\textit{(+40.2\%)}}&		\textbf{\textit{(+43.9\%)}}&		\textit{(+4.0\%)}&		\textit{(+5.0\%)}	 \\ \midrule
\multirow{3}{*}{Encoding}&\multicolumn{2}{g}{Tamil}&\multicolumn{2}{c}{Uyghur}&\multicolumn{2}{g}{Wolof}\\

&\tiny{UAS}&\tiny{LAS}&\tiny{UAS}&\tiny{LAS}&\tiny{UAS}&\tiny{LAS}\\
\hline
rel-PoS &76.5	\textit{(+31.9\%)}&	68.7	\textit{(+37.9\%)}&	74.3	\textit{(+4.0\%)}&	61.1	\textit{(+5.7\%)}&	86.0	\textit{(+13.0\%)}&	81.8	\textit{(+17.5\%)} \\
Brackets  & 73.6	\textit{(+28.5\%)}&	66.8	\textit{(+36.3\%)}&	74.7	\textit{(+3.7\%)}&	61.7	\textit{(+5.5\%)}&	85.3	\textit{(+13.1\%)}&	80.7	\textit{(+17.2\%)}	 \\
Standard & 73.3	\textit{(+28.2\%)}&	66.7	\textit{(+37.2\%)}&	74.4	\textit{(+4.0\%)}&	61.7	\textit{(+6.0\%)}&	80.6	\textit{(+11.9\%)}&	76.5	\textit{(+16.6\%)}	 \\
Eager  &  74.6	\textit{(+32.8\%)}&	67.6	\textit{(+40.1\%)}&	74.3	\textit{(+3.3\%)}&	61.3	\textit{(+4.3\%)}&	82.5	\textit{(+12.2\%)}&	78.1	\textit{(+16.4\%)}	 \\
Hybrid  & 72.8	\textit{(+27.7\%)}&	66.0	\textit{(+35.5\%)}&	73.6	\textit{(+3.2\%)}&	61.5	\textit{(+5.0\%)}&	80.8	\textit{(+11.9\%)}&	76.5	\textit{(+16.5\%)}	 \\
Covington & 66.6	\textit{(+27.4\%)}&	59.9	\textit{(+33.8\%)}&	68.7	\textit{(+3.5\%)}&	56.5	\textit{(+5.7\%)}&	78.1	\textit{(+11.7\%)}&	74.2	\textit{(+x\%)}	 \\ \midrule

\textit{avg}  &\textbf{\textit{(+29.4\%)}}&		\textbf{\textit{(+36.8\%)}}&		\textit{(+3.6\%)}&		\textit{(+5.4\%)}&		\textit{(+12.3\%)}&		\textit{(+16.7\%)}	 \\ \bottomrule
\end{tabular}}
\caption{Performance of the models trained with gold segmentation, tokenization and PoS tags on the test sets, reported in \textsc{UAS} and \textsc{LAS}, with the percentage increase with respect to the results obtained with the predicted setup. The treebanks with the three highest averaged \% of improvement in bold. }
\label{tab:gold}
\end{table*}

\begin{table}[hptb!]
\centering
\begin{small}
\renewcommand{\arraystretch}{0.85}
\begin{adjustbox}{max width=1\columnwidth}
\begin{tabular}{@{}lccc}
\toprule
Language & Words (\%) & Sentences (\%) & UPoS (\%)  \\ \midrule
\rowcolor[HTML]{EFEFEF} 
\begin{tabular}[c]{@{}l@{}}Ancient\\ Greek\end{tabular} & 100.0 & 98.9 & 82.2 \\
Chinese & 90.0 & 99.1 & 84.0  \\
\rowcolor[HTML]{EFEFEF} 
English & 99.0 & 76.3 & 93.5  \\
Finnish & 99.7 & 88.6 & 94.3  \\
\rowcolor[HTML]{EFEFEF} 
Hebrew & 85.0 & 99.4 & 80.5  \\
Russian & 99.5 & 96.2 & 95.0  \\
\rowcolor[HTML]{EFEFEF} 
Tamil & 94.5 & 97.5 & 81.3  \\
Uyghur & 99.7 & 82.9 & 87.9  \\
\rowcolor[HTML]{EFEFEF} 
Wolof & 99.2 & 92.0 & 91.7  \\ \bottomrule
\end{tabular}
\end{adjustbox}
\end{small}
\caption{Prediction accuracy of UDpipe 1 on UDv2.4 treebanks.}
\label{tab:udpipe}
\end{table}

\end{document}